\documentclass[10pt,twocolumn,letterpaper]{article}

\usepackage{cvpr}              

\usepackage{times}
\usepackage{epsfig}
\usepackage{graphicx}
\usepackage{amsmath}
\usepackage{amssymb}
\usepackage{subcaption}
\usepackage{multirow}
\usepackage{placeins}
\usepackage{floatrow}
\usepackage{asymptote}
\usepackage{subcaption}
\usepackage{mathtools}

\usepackage{tikz}
\usetikzlibrary{fit}
\usetikzlibrary{positioning}
\usetikzlibrary{calc}
\usetikzlibrary{decorations.pathreplacing,calligraphy}
\usetikzlibrary{shapes.geometric}

\definecolor{cvprblue}{rgb}{0.21,0.49,0.74}
\usepackage[pagebackref,breaklinks,colorlinks,citecolor=cvprblue,hypertexnames=false]{hyperref}

\def\paperID{18} 
\def\confName{CVPR}
\def\confYear{2024}

\title{Lift-Attend-Splat: Bird’s-eye-view camera-lidar fusion using transformers}

\author{
    James Gunn\thanks{Equal contribution} \qquad
    Zygmunt Lenyk\footnotemark[1] \qquad
    Anuj Sharma \qquad
    Andrea Donati \\
    Alexandru Buburuzan \qquad
    John Redford \qquad
    Romain Mueller \\
    FiveAI \\
     {\tt\small <first>.<last>@five.ai}
}

\begin{document}
\maketitle

\begin{abstract}
Combining complementary sensor modalities is crucial to providing robust perception for safety-critical robotics applications such as autonomous driving (AD).
Recent state-of-the-art camera-lidar fusion methods for AD rely on monocular depth estimation which is a notoriously difficult task compared to using depth information from the lidar directly.
Here, we find that this approach does not leverage depth as expected and show that naively improving depth estimation does not lead to improvements in object detection performance. Strikingly, we also find that removing depth estimation altogether does not degrade object detection performance substantially, suggesting that relying on monocular depth could be an unnecessary architectural bottleneck during camera-lidar fusion.
In this work, we introduce a novel fusion method that bypasses monocular depth estimation altogether and instead selects and fuses camera and lidar features in a bird's-eye-view grid using a simple attention mechanism.
We show that our model can modulate its use of camera features based on the availability of lidar features and that it yields better 3D object detection on the nuScenes dataset than baselines relying on monocular depth estimation.
\end{abstract}


\section{Introduction}

Integrating information from different modalities efficiently and effectively is especially important in safety-critical applications such as autonomous driving, where different sensor modalities are complementary and combining them adequately is crucial to guarantee safety.
For example, cameras capture rich semantic information of objects up to far away distances, while lidars provide extremely accurate depth information but are sparse at large distances.
For this reason, many modern self-driving platforms have a large number of different sensors which must be combined in order to provide accurate and reliable perception of the surrounding scene and allow safe deployment of these vehicles in the real world.

Multimodal sensor fusion --- learning a unified representation of a scene derived from multiple sensors --- offers a plausible solution to this problem.
However, training such multimodal models can be challenging, especially when modalities are as different as cameras (RGB images) and lidars (3D point clouds).
For instance, it is known that different modalities overfit and generalise at different rates~\cite{wang2020makes} and that training all modalities jointly can lead to underutilisation of the weaker modalities and even to inferior results compared to unimodal models in some situations~\cite{parida2020coordinated}.

In the context of autonomous driving, many of the recent state-of-the-art methods for camera-lidar fusion~\cite{liu2022bevfusion,liang2022bevfusion,ea-bev} are based on the Lift-Splat (LS) paradigm~\cite{philion2020lift}\footnote{The ``shoot'' component of ``Lift, Splat, Shoot''~\cite{philion2020lift} relates to trajectory prediction and is not considered here.}.
In this approach, the camera features are projected in bird's-eye-view (BEV) --- or top-down space --- using monocular depth before being fused with the lidar features.
As a result, the location of the camera features in BEV is highly dependent on the quality of the monocular depth prediction and it has been argued that its accuracy is critical~\cite{liang2022bevfusion,ea-bev}.
In this work, we reconsider these claims and show that the monocular depth prediction inside these models is of poor quality and cannot account for their success.
In particular, we present results showing that methods based on Lift-Splat perform equally well when the monocular depth prediction is replaced by direct depth estimation from the lidar point cloud or removed completely.
This leads us to suggest that relying on monocular depth when fusing camera and lidar features is an unnecessary architectural bottleneck and that Lift-Splat could be replaced by a more effective projection.

We introduce a novel approach for camera-lidar fusion called ``Lift-Attend-Splat'' that bypasses monocular depth estimation altogether and instead selects and fuses camera and lidar features in BEV using a simple transformer.
We present evidence that our method shows better camera utilisation compared to the methods based on monocular depth estimation and that it improves object detection performance.
Our contributions are as follows:
\begin{itemize}
    \item We show that camera-lidar fusion methods based on the Lift-Splat paradigm are not leveraging depth as expected. In particular, we show that they perform equivalently or better if monocular depth prediction is removed completely.
    \item We introduce a novel camera-lidar fusion method that fuses camera and lidar features in BEV using a simple attention mechanism.
    We show that it leads to better camera utilisation and improves 3D object detection compared to models based on the Lift-Splat paradigm.
\end{itemize}

\section{Related work}

\paragraph{3D object detection for autonomous driving}
Most 3D object detection benchmarks are dominated by methods using lidar point clouds due to their highly accurate range measurement allowing for better placement of objects in 3D compared to methods using cameras or radars only.
Deep learning methods for classification on point clouds were pioneered in the seminal works of~\cite{qi2017pointnet,qi2017pointnet++} and early works have been applying similar ideas to 3D object detection~\cite{qi2018frustum,shi2019pointrcnn}.
A more recent family of methods is based on direct voxelisation of the 3D space~\cite{zhou2018voxelnet,yan2018second} or compression of the lidar representation along the z-direction into ``pillars''~\cite{lang2019pointpillars,yang2018pixor}.
These approaches have been very successful and are the basis of many follow-up works~\cite{yin2021centerbased,huang2022rethinking,koh2023mgtanet}.

The task of 3D object detection has also been tackled from multiple cameras alone.
Early works have mostly been based on various two-stage approaches~\cite{chen2017multi,qi2018frustum,ku2018joint,wang2019frustum}, while recent methods have been leveraging monocular depth estimation directly~\cite{7780605,ku2019monocular,qian2020end}.
This task is difficult when lidar is absent because 3D information must be estimated using images only, which is a challenging problem.
However, recent works have shown impressive performance by borrowing ideas from lidar detection pipelines~\cite{huang2021bevdet,chong2022monodistill,guo2021liga}, by improving position embeddings~\cite{liu2022petr} and 3D queries~\cite{jiang2023far3d}, as well as by leveraging temporal aggregation~\cite{zong2023temporal,han2023exploring,liu2023petrv2,wang2023exploring,lin2022sparse4d,li2022bevformer} or 2D semantic segmentation~\cite{zhang2023sa}.

\paragraph{Camera-lidar fusion}
Perception quality can be improved by jointly leveraging cameras and lidars when available.
Recent fusion methods can be broadly classified into three categories: point decoration methods, methods that leverage task-specific object queries and architectures, and projection based methods.
Point decoration methods augment the lidar point cloud using semantic segmentation data~\cite{vora2020pointpainting,xu2021fusionpainting}, camera features~\cite{wang2021pointaugmenting}, or even create new 3D points using object detections in the image plane~\cite{yin2021multimodal}.
Such methods are relatively easy to implement but suffer from the fact that they require lidar points to fuse camera features.
TransFusion~\cite{bai2022transfusion} is a recent example of a method that leverages task-specific object queries generated using the lidar point cloud.
Final detections are made directly without explicit projection of camera features into BEV space. Similar methods utilising two-way modality interactions perform even better~\cite{yang2022deepinteraction, yan2023cross}.
Fusion can also be performed earlier in the model, for example at the level of the 3D voxels~\cite{chen2022autoalign,chen2022focal} or lidar features~\cite{li2022deepfusion}, or by sharing information between the camera and lidar backbones~\cite{liang2018deep, huang2020epnet, piergiovanni20214d}. MSMDFusion~\cite{jiao2023msmdfusion} fuses camera and lidar features at multiple scales using lidar points to estimate the 3D position of camera features, while UniTR~\cite{wang2023unitr} pre-assigns depth to each camera feature. FUTR3D~\cite{chen2023futr3d} fuses features by selecting modality-agnostic 3D reference points.
Finally, projection-based methods project camera features into 3D before fusing them with the lidar (see below).

\paragraph{Projection based methods}
Of most interest to us are camera-lidar fusion methods based on projecting camera features into 3D.
Many recent state-of-the-art camera-lidar fusion methods~\cite{liu2022bevfusion,liang2022bevfusion,ea-bev} project camera features in 3D using monocular depth estimation even though depth information is available from the lidar. It has also been shown that the projection method is less important than other aspects of training in the camera-only setting~\cite{harley2023simple}.
An alternative approach is to project camera features directly into BEV space using the known correspondence between lidar points and camera features~\cite{drews2022deepfusion, wang2018fusing, li2022deepfusion}.
However, the sparsity of the lidar point cloud can limit which camera features are projected, as described in~\cite{liu2022bevfusion}. 
Finally, learning to project camera features in BEV without explicit depth can be achieved when lidar is absent using a transformer, as shown in~\cite{saha2022translating,li2022bevformer}.
Here, we extend this line of work to the case of camera-lidar fusion and leverage cross-attention to generate a dense BEV grid of fused lidar features.

\begin{figure*}
\begin{center}
\begin{minipage}{0.48\textwidth}
  \begin{subfigure}{0.48\textwidth}
    \centering
    \caption*{Camera}
    \includegraphics[width=\linewidth]{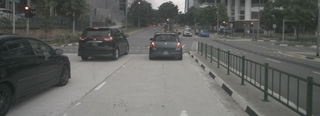}
  \end{subfigure}
  \begin{subfigure}{0.48\textwidth}
    \centering
    \caption*{Lidar}
    \includegraphics[width=\linewidth]{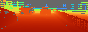}
  \end{subfigure}
  \\
  \begin{subfigure}{0.48\textwidth}
    \centering
    \includegraphics[width=\linewidth]{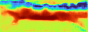}
    \caption*{BEVFusion~\cite{liu2022bevfusion}}
  \end{subfigure}
  \begin{subfigure}{0.48\textwidth}
    \centering
    \includegraphics[width=\linewidth]{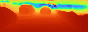}
    \caption*{BEVFusion~\cite{liu2022bevfusion}~w/ $\lambda = 1$}
  \end{subfigure}
\end{minipage}
\hspace{1cm}
\begin{tabular}{lcc|c}
~ & Abs. Rel.~$\downarrow$ & RMSE~$\downarrow$ & mAP~$\uparrow$ \\
\hline\hline
BEVFusion~\cite{liu2022bevfusion} & 2.75 & 17.40 & \textbf{68.5} \\
\hline
\multicolumn{3}{l}{BEVFusion~\cite{liu2022bevfusion} w/ \cref{eq:depth sup loss}:} \\ 
\qquad $\lambda = 0$ & 2.83 & 18.54 & 68.4 \\
\qquad $\lambda = 0.01$ & 0.76 & 8.09 & 68.0 \\
\qquad $\lambda = 1$ & 0.22 & 4.77 & 68.1 \\
\qquad $\lambda = 100$ & 0.16 & 4.55 & 64.6 \\
\hline
Lidar & 0.04 & 0.29 & 68.4 \\
Pretrained & 0.64 & 7.87 & 67.4 \\
Uniform depth & -- & -- & \textbf{68.5} \\
\end{tabular}
\end{center}
\caption{\label{fig:depth quality}
Impact of the quality of the monocular depth prediction on the object detection performance of BEVFusion~\cite{liu2022bevfusion} on the nuScenes validation set.
We compare BEVFusion and four different variants: adding depth supervision using \cref{eq:depth sup loss} with various weights $\lambda$, using lidar depth maps instead of monocular depth estimation (lidar), using a pretrained and frozen depth classifier (pretrained), and finally removing depth estimation altogether by projecting camera features at all depths uniformly using \cref{eq:swiftblat} (uniform depth).
In our experiments, more accurate depth does not translate to better detection performance and the original model is on-par with using the lidar points directly as a source of depth.
Equivalent detection performance was achieved using the \emph{uniform depth} model, clearly indicating that accurate monocular depth is not necessary for BEVFusion~\cite{liu2022bevfusion} to achieve its performance, see main text and~\cref{sec:suppl:depth results} for details.
}
\end{figure*}

\section{Monocular depth prediction in Lift-Splat} \label{sec:depth}

Recent camera-lidar fusion methods based on the Lift-Splat paradigm~\cite{liu2022bevfusion,liang2022bevfusion} learn a unified representation in the form of a BEV grid by projecting camera features in BEV space using an estimated depth distribution as  
\begin{equation} \label{eq:Lift-Splat}
    \text{Proj}_{\text{Lift-Splat}}
    = \text{Splat}\left({F'}^{\text{cam}} \otimes D \right),
\end{equation}
where ${F'}^{\text{cam}} \in \mathbb{R}^{C_c' \times H \times W }$ is a context vector obtained from the camera features $F^{\text{cam}} \in \mathbb{R}^{C_c \times H \times W }$, $D \in \mathbb{R}^{N_D \times H \times W}$ is a normalised distribution over predetermined depth bins and $\text{Splat}$ denotes the operation of projecting each point downwards into the $z=0$ plane, see~\cite{liu2022bevfusion,liang2022bevfusion,philion2020lift} for details.
The resulting feature map is then merged with the lidar features using concatenation~\cite{liu2022bevfusion} or gated attention~\cite{liang2022bevfusion}.
In this paradigm, the monocular depth distribution prediction is learned indirectly from the downstream task without explicit depth supervision.

\paragraph{Lift-Splat depth prediction is generally poor}
We analyse the quality of the depth distribution predicted by BEVFusion~\cite{liu2022bevfusion} by comparing its mean value to lidar depth maps, both qualitatively and quantitatively using the absolute relative (Abs. Rel.) and root mean squared errors (RMSE)~\cite{eigen2014depth,li2022bevdepth}.
As shown on \cref{fig:depth quality}, the mean depth prediction does not accurately reflect the structure of the scene and is markedly different from the lidar depth map which suggests that monocular depth is not leveraged as expected in~\cite{liu2022bevfusion}.
See \cref{sec:suppl:depth viz,sec:suppl:depth results} for details.

\paragraph{Improving depth prediction does not improve detection performance}
We next investigate whether improving the depth prediction quality can boost object detection performance.
To do so, we retrain the model from~\cite{liu2022bevfusion} with the following loss:
\begin{equation}
    \label{eq:depth sup loss}
    L_{\text{total}} = L_{\text{sup}}+\lambda L_{\text{depth}},
\end{equation}
where $L_{\text{sup}}$ is the original 3D object detection loss and $L_{\text{depth}}$ is a simple cross-entropy loss for the depth estimation that uses one-hot encoded lidar depth as a target, see \cref{sec:suppl:depth supervision} for more details.
By changing the hyper-parameter $\lambda$, we can control the quality of the depth prediction and explore how it impacts detection performance. 
In \cref{fig:depth quality}, we see that while depth supervision indeed leads to much more accurate depth maps both visually and quantitatively, detection performance --- measured using mean average precision (mAP) --- degrades from the baseline as the weight of the depth supervision is increased.
This suggests that the method is unable to take advantage of more accurate depth prediction.
Since training on the multi-task loss \cref{eq:depth loss} is likely to degrade object detection performance at high values of $\lambda$, we also experiment with two more variants: (i) pretraining the depth supervision module separately and (ii) using the lidar point cloud directly to bypass the depth supervision module altogether.
Pretraining leads to more accurate depth prediction but degrades detection performance relative to the baseline, while using the lidar directly does not change the detection performance compared to the baseline, even though all depth metrics are close to zero\footnote{They are not exactly zero because of the depth quantisation introduced by the one-hot encoding of the lidar depth, see \cref{sec:suppl:gtdepth}.}.

\paragraph{Removing depth prediction altogether does not affect object detection performance}

The results above lead us to hypothesise that accurate monocular depth is not leveraged in camera-lidar fusion methods based on the Lift-Splat projection.
To test this, we remove the monocular depth prediction completely and replace the projection~\eqref{eq:Lift-Splat} by
\begin{equation} \label{eq:swiftblat}
    \text{Proj}_{\text{no-depth}}
    = \text{Splat}\left({F'}^{\text{cam}} \otimes 1 \right),
\end{equation}
where we denote by $1$ the tensor of the same shape as $D$ with all entries equal to $1$.
This projects the camera features to all depths uniformly.
Strikingly, we see in \cref{fig:depth quality} (right) that \emph{removing monocular depth estimation does not lead to a degradation in object detection performance}, suggesting that accurate depth estimation is not a key component of this method.
We hypothesise that the importance of monocular depth is greatly diminished when lidar features are available since lidar is a much more precise source of depth information and that the model is able to easily suppress camera features projected at the wrong location.
This suggests that relying on monocular depth estimation could be an unnecessary architectural bottleneck and lead to underutilisation of the camera.

\section{Camera-lidar fusion without monocular depth estimation} \label{sec:method}
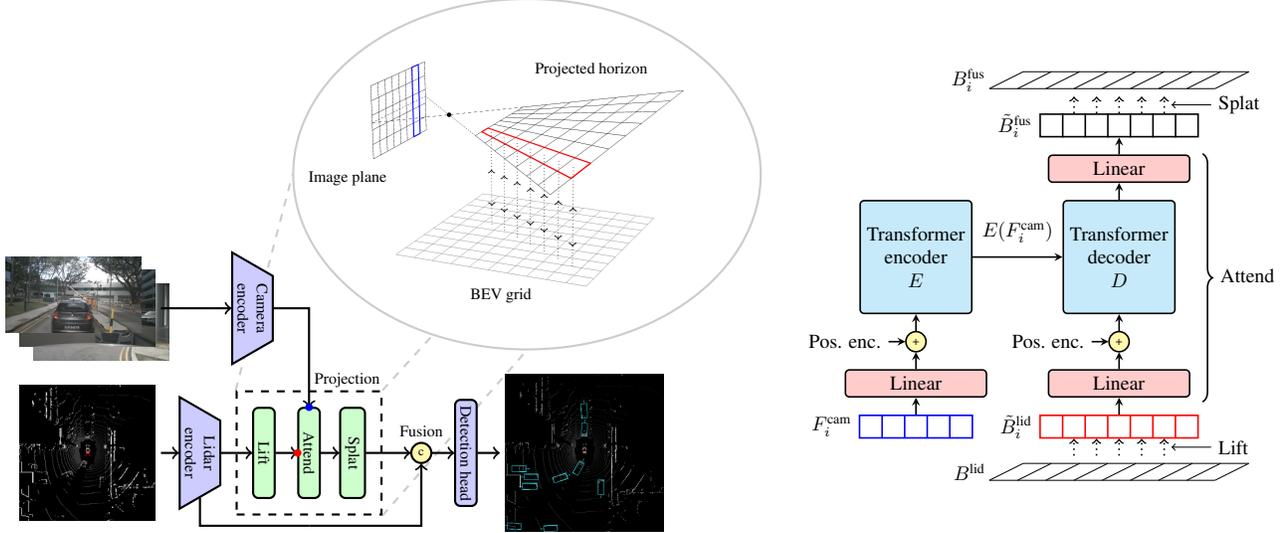
\begin{figure*}[!ht]
\centering
\begin{subfigure}{0.58\textwidth}
\begin{tikzpicture}[scale=0.6]
\pgfdeclarelayer{bg}    
\pgfsetlayers{bg,main}
\tikzset{every node/.style={thick, transform shape}}
\tikzset{lin/.style={draw,thick, rounded corners=2, fill=green!20}}
\tikzset{trf/.style={draw,thick, rounded corners=2, fill=green!20}}
\tikzset{mrg/.style={draw,thick, rounded corners=2, fill=green!20}}
\tikzset{det/.style={draw,thick, rounded corners=2, fill=blue!20}}
\tikzset{enc/.style={draw,thick, trapezium, rotate=270, minimum width=1.5cm, fill=blue!20}}
\tikzset{-/.style={thick}}
\tikzstyle{fitbox}=[draw, inner sep=6pt, dashed]
\node (lift) [lin, minimum width=2cm, rotate=-90, anchor=north] {Lift};
\node (attend) [trf, right =of lift.north, minimum width=2cm, rotate=-90, anchor=north] {Attend};
\node (splat) [lin, right =of attend.north, minimum width=2cm, rotate=-90, anchor=north] {Splat};
\node (merge) [right=of splat.north, circle, draw, scale=0.8, fill=yellow!40] {c};
\node [above=0cm of merge] {Fusion};
\node (projbox) [fit=(lift) (attend) (splat), fitbox] {};
\node (projtext) [anchor=south west, yshift=-0.05cm, thin] at (projbox.north) {Projection};
\node (lid encoder) [enc, left =0.7cm of lift.south, anchor=north, text width=1.2cm, align=center] {Lidar encoder};
\node (cam encoder) [enc, above =1.2cm of lift.south west, anchor=east, text width=1.2cm, align=center] {Camera encoder};
\node (lidar) [left=0.4cm of lid encoder.south] {\includegraphics[width=3cm]{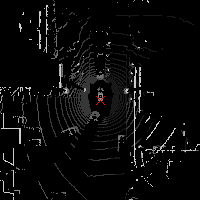}};
\node at (lidar |- cam encoder) [xshift=0.3cm, yshift=-0.3cm] {\includegraphics[width=3cm]{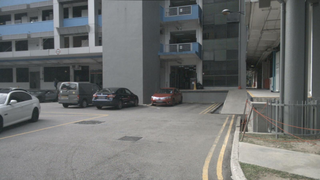}};
\node (images) at (lidar |- cam encoder) {\includegraphics[width=3cm]{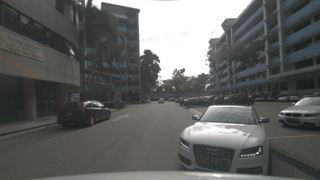}};
\node at (lidar |- cam encoder) [xshift=-0.3cm, yshift=0.3cm] {\includegraphics[width=3cm]{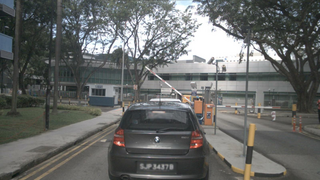}};
\node (det head) [det, right=0.5cm of merge, rotate=-90, anchor=south] {Detection head};
\node (out) [right =0.5cm of det head.north] {\includegraphics[width=3.5cm]{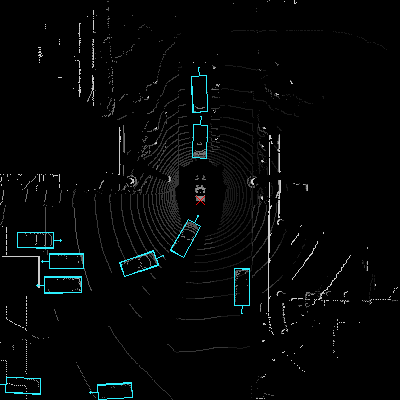}};
\node (geometry) [draw=gray!40, fill=white, ellipse, above=0.4cm of out.north west, minimum width=1cm, minimum height=1cm, xshift=0.6cm] {
    \begin{subfigure}{0.7\textwidth}
    \includegraphics[width=\textwidth]{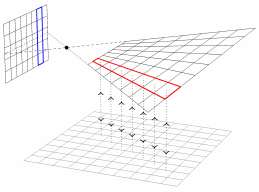}
    \end{subfigure}
};
\node [yshift=6.1cm, xshift=1.6cm] {Image plane};
\node [yshift=8.5cm, xshift=7cm] {Projected horizon};
\node [yshift=3.5cm, xshift=5cm] {BEV grid};
\draw [->,thick] (lift.north) -- (attend.south);
\draw [->,thick] (attend.north) -- (splat.south);
\draw [->,thick] (splat.north) -- (merge);
\draw [->,thick] (images.east) -- (cam encoder.south);
\draw [->,thick] (lidar.east) -- (lid encoder.south);
\draw [->,thick] (cam encoder.north) -| (attend.west);
\draw [->,thick] (lid encoder) -- (lift);
\draw [->,thick] (merge) -- (det head);
\draw [->,thick] (det head) -- (out.west |- det head);
\begin{pgfonlayer}{bg}    
    \draw [-, gray!40, dashed] (projbox.north west) -- (geometry.west);
    \draw [-, gray!40, dashed] (projbox.south east) -- (geometry.south);
    \draw [-, gray!40, dashed] (projbox.north east) -- (geometry.north east);
\end{pgfonlayer}
\node at (attend.west) [circle,fill,inner sep=1.7pt,blue]{};
\node at (attend.south) [circle,fill,inner sep=1.7pt,red]{};
\node (spacer) [below=1.25cm of merge.south] {};
\coordinate (elbow) at (attend |- spacer) {}; 
\draw [-,thick] (lid encoder.east) |- (elbow);
\draw [->,thick] (elbow) -| (merge);
\end{tikzpicture}
\end{subfigure}
\begin{subfigure}{0.4\textwidth}
\begin{tikzpicture}[align=center,node distance=0.3cm,scale=0.8,transform canvas={scale=0.75}]
\begin{scope}[shift={(2cm,2.2cm)}]
\tikzset{every node/.style={thick}}
\tikzset{lin/.style={draw,thick,rounded corners=2,minimum width=2.5cm,fill=pink!80}}
\tikzset{trf/.style={draw,thick,rounded corners=2,minimum height=2cm,fill=cyan!20}}
\tikzset{-/.style={thick}}
\node[fit={(0,0) (2.5,0.5)}, inner sep=0pt, draw=blue, thick] (cam) {};
\draw[-,thick,blue] (0.5,0) -- (0.5,0.5);
\draw[-,thick,blue] (1.0,0) -- (1.0,0.5);
\draw[-,thick,blue] (1.5,0) -- (1.5,0.5);
\draw[-,thick,blue] (2.0,0) -- (2.0,0.5);
\node[left=0.0cm of cam] {$F^{\text{cam}}_i$};
\node[lin,above=of cam] (camlin) {Linear};
\draw[->,thick,>=stealth] (cam.north -| camlin) -- (camlin);
\node[above=of camlin,circle,draw,scale=0.6,fill=yellow!40] (camplus) {+};
\draw[->,thick,>=stealth] (camlin) -- (camplus);
\node[left=of camplus] (campos) {Pos. enc.};
\draw[->,thick,>=stealth] (campos) -- (camplus);
\node[trf, above=of camplus,minimum height=2cm] (enc) {Transformer \\encoder\\$E$};
\draw[->,thick,>=stealth] (camplus) -- (enc);
\node[fit={(4.0,0) (7.5,0.5)},inner sep=0pt,draw=red,thick] (lidar) {};
\draw[-,thick,red] (4.5,0) -- (4.5,0.5);
\draw[-,thick,red] (5.0,0) -- (5.0,0.5);
\draw[-,thick,red] (5.5,0) -- (5.5,0.5);
\draw[-,thick,red] (6.0,0) -- (6.0,0.5);
\draw[-,thick,red] (6.5,0) -- (6.5,0.5);
\draw[-,thick,red] (7.0,0) -- (7.0,0.5);
\node[left=0.0cm of lidar] {$\tilde B^{\text{lid}}_i$};
\node[below=of lidar] (bev) {
\begin{tikzpicture}
\draw (0,0) -- (4,0);
\draw (0.6,0.3) -- (4.6,0.3);
\foreach \i in {0,...,8} {
    \draw (\i/2,0) -- (\i/2+0.6,0.3);
}\end{tikzpicture}};
\node[left=-0.2cm of bev] {$B^{\text{lid}}$};
\draw[->,dotted,thick] (bev.north) -- (lidar);
\draw[->,dotted,thick] ([xshift=0.5cm]bev.north) -- ([xshift=0.5cm]lidar.south);
\draw[->,dotted,thick] ([xshift=1.0cm]bev.north) -- ([xshift=1.0cm]lidar.south);
\draw[->,dotted,thick] ([xshift=-0.5cm]bev.north) -- ([xshift=-0.5cm]lidar.south);
\draw[->,dotted,thick] ([xshift=-1.0cm]bev.north) -- ([xshift=-1.0cm]lidar.south);
\node (last arrow1) at ($([xshift=1.0cm]bev.north)!0.5!([xshift=1.0cm]lidar.south)$) {};
\node[right=0.7cm of last arrow1] (lift) {Lift};
\draw[->] (lift) -- (last arrow1);
\node[lin, above=of lidar] (lidlin) {Linear};
\draw[->,thick,>=stealth] (lidar.north -| lidlin) -- (lidlin);
\node[above=of lidlin,circle,draw,scale=0.6,fill=yellow!40] (lidplus) {+};
\draw[->,thick,>=stealth] (lidlin) -- (lidplus);
\node[left=of lidplus] (lidpos) {Pos. enc.};
\draw[->,thick,>=stealth] (lidpos) -- (lidplus);
\node[trf,above=of lidplus] (dec) {Transformer \\decoder\\$D$};
\draw[->,thick,>=stealth] (lidplus) -- (dec);
\draw[->,thick,>=stealth] (enc.east) -- (dec.west);
\node (encdec) at ($(enc.east)!0.5!(dec.west)$) {};
\node[above=0.0cm of encdec] {$E(F_i^{\text{cam}})$};
\node[lin,above=of dec] (lidlin2) {Linear};
\draw[->,thick,>=stealth] (dec.north) -- (lidlin2);
\node[fit={(0.0,0) (3.5,0.5)}, above=of lidlin2, inner sep=0pt, draw=black, thick] (result) {};
\draw[-,thick] ([xshift=0.25cm]result.north) -- ([xshift=0.25cm]result.south);
\draw[-,thick] ([xshift=0.75cm]result.north) -- ([xshift=0.75cm]result.south);
\draw[-,thick] ([xshift=1.25cm]result.north) -- ([xshift=1.25cm]result.south);
\draw[-,thick] ([xshift=-0.25cm]result.north) -- ([xshift=-0.25cm]result.south);
\draw[-,thick] ([xshift=-0.75cm]result.north) -- ([xshift=-0.75cm]result.south);
\draw[-,thick] ([xshift=-1.25cm]result.north) -- ([xshift=-1.25cm]result.south);
\draw[->,thick,>=stealth] (lidlin2) -- (result);
\node[left=0.0cm of result] {$\tilde B^{\text{fus}}_i$};
\node[above=of result] (bev2) {
\begin{tikzpicture}
\draw (0,0) -- (4,0);
\draw (0.6,0.3) -- (4.6,0.3);
\foreach \i in {0,...,8} {
    \draw (\i/2,0) -- (\i/2+0.6,0.3);
}\end{tikzpicture}};
\node[left=-0.2cm of bev2] {$B^{\text{fus}}_i$};
\draw[->,dotted,thick] (result.north) -- (bev2.south);
\draw[->,dotted,thick] ([xshift=0.5cm]result.north) -- ([xshift=0.5cm]bev2.south);
\draw[->,dotted,thick] ([xshift=1.0cm]result.north) -- ([xshift=1.0cm]bev2.south);
\draw[->,dotted,thick] ([xshift=-0.5cm]result.north) -- ([xshift=-0.5cm]bev2.south);
\draw[->,dotted,thick] ([xshift=-1.0cm]result.north) -- ([xshift=-1.0cm]bev2.south);
\node (last arrow2) at ($([xshift=1.0cm]result.north)!0.5!([xshift=1.0cm]bev2.south)$) {};
\node[right=0.7cm of last arrow2] (proj){Splat};
\draw[->] (proj) -- (last arrow2);
\draw[decorate, decoration = {brace, amplitude=5pt}, thick] (7.6,6.3) -- (7.6,0.8);
\node[fit={(8,3) (8,3.9)}] {Attend};
\end{scope}
\end{tikzpicture}
\end{subfigure}
\caption{\label{fig:transformer}
Lift-Attend-Splat camera-lidar fusion architecture.
(left) Overall architecture: features from the camera and lidar backbones are fused together and merged before being passed to a detection head.
(inset) Geometry of our 3D projection: the ``Lift'' step embeds the lidar BEV features into the projected horizon by lifting the lidar features along the z-direction using bilinear sampling.
The ``Splat'' step corresponds to the inverse transformation in that it projects features from the projected horizon back onto the BEV grid using bilinear sampling, again along the z-direction.
(right) Details of the projection module: the ``Attend'' step in our method lets the lifted lidar features $\tilde B^{\text{lid}}_i$ attend to the camera features $F^{\text{cam}}_i$ in the corresponding column using a simple encoder-decoder transformer architecture to produce fused features $D(\tilde B_i^{\text{lid}}, E(F^{\text{cam}}_i))$ in frustum space.
}
\end{figure*}

In this section, we present a camera-lidar fusion method that bypasses monocular depth estimation altogether and instead fuses camera and lidar features in bird's-eye-view using a simple transformer~\cite{vaswani2017attention}.
A naive application of the transformer architecture to the problem of camera-lidar fusion is difficult, however, due to the large number of camera and lidar features and the quadratic nature of attention.
As shown in~\cite{saha2022translating}, it is possible to use the geometry of the problem to drastically restrict the scope of the attention when projecting camera features in BEV, since camera features should only contribute to locations along their corresponding rays.
We adapt this idea to the case of camera-lidar fusion and introduce a simple method that uses cross-attention between columns in the camera plane and polar rays in the lidar BEV grid.
Instead of predicting monocular depth, cross-attention learns which camera features are the most salient given context provided by the lidar features along its ray.

Except for the projection of the camera features in BEV, our model shares a similar overall architecture to methods based on the Lift-Splat paradigm~\cite{liu2022bevfusion,liang2022bevfusion,ea-bev} and is depicted on \cref{fig:transformer} left.
It consists of the following modules: the camera and lidar backbones which produce features for each modality independently, a projection and fusion module that embeds the camera features into BEV and fuses them with the lidar, and finally a detection head.
When considering object detection, the final output of the model is the property of objects in the scene represented as 3D bounding boxes with position, dimension, orientation, velocity and classification information.
In what follows we explain in detail the architecture of our projection and fusion modules.

\paragraph{Projected horizon}
For each camera, we consider the horizontal line passing through the centre of the image and the plane corresponding to its projection in 3D.
We call this plane the \emph{projected horizon} of the camera.
It can easily be described using homogeneous coordinates as the set of points $\mathbf x \in \mathbb R^4$ for which there exists a $u \in \mathbb R$ such that
\begin{equation}
    \mathbf C \mathbf x \sim (u, h/2, 1),
\end{equation}
where $\mathbf C$ is the $3 \times 4$ camera projection matrix (intrinsic and extrinsic), and $h$ is the height of the image.
Note that this plane is not in general parallel to the BEV grid, its relative orientation being defined by the camera's extrinsic parameters.
We define a regular grid on each camera's projected horizon that is aligned with the 2D grid of features in its image plane by tracing out rays from the intersection of the horizontal line with the edges of the feature columns in the image plane, and then separating these rays into a set of predetermined depth bins (similarly to~\cite{liang2022bevfusion}).
Features on this grid can be represented by a matrix $G \in \mathbb{R}^{N_D \times W}$, where each row corresponds to a specific column in the camera feature map $F^{\text{cam}} \in \mathbb{R}^{H \times W \times C}$.
The geometry of a projected horizon can be seen in \cref{fig:transformer} (left, inset).
The projected horizon allows for a consistent definition of depth between differently pitched cameras.

\paragraph{Correspondence between projected horizons and BEV grid}
We can easily define a correspondence between points on a projected horizon and points on the BEV plane by projecting them along the z-direction in 3D space.
As cameras are in general tilted with respect to the ground, this correspondence depends on each camera's extrinsic parameters.
We transfer lidar features from the BEV grid to a camera's projected horizon through bi-linear sampling of the BEV grid at the locations of the down-projected cell-centers of the projected horizon.
We call this process ``lifting'' and denote it as $\text{Lift}_i$ for the projected horizon of camera $i$.
Similarly, camera features can be transferred in the opposite direction, from a projected horizon to the BEV grid, by bi-linearly sampling the projected horizon at the locations of the projected cell-centers of the BEV grid.
We denote this operation as $\text{Splat}_i$, similarly to \cite{philion2020lift,liu2022bevfusion,liang2022bevfusion}.
Fusion of lidar features with splatted camera features takes place in BEV space, as is common~\cite{liu2022bevfusion, liang2022bevfusion}.

\paragraph{Lift-Attend-Splat}
Our projection module is depicted in \cref{fig:transformer} (right) and can be broken down into three simple steps:
(i) we first \emph{lift} the BEV lidar features $B^{\text{lid}}$ onto the projected horizon of camera $i$, producing ``lifted'' lidar features $\tilde B^{\text{lid}}_i$,
(ii) we then let the ``lifted'' lidar features \emph{attend} to the camera features in the corresponding column using a simple transformer encoder-decoder, producing fused features $\tilde B^{\text{fus}}_i$ on the $i^{\text{th}}$ projected horizon, and finally
(iii) we \emph{splat} these features back onto the BEV grid to produce $B^{\text{fus}}_i$.
During the attend step, the camera features in each column are encoded by a transformer encoder $E$ and passed as keys and values to a transformer decoder $D$ which uses the frustum lidar features as queries.
The result of these three steps can be written as
\begin{equation}
        B^{\text{fus}}_i = \text{Splat}_i\left(D\left(\text{Lift}_i \left(B^{\text{lid}}\right), E\left(F_i^{\text{cam}}\right)\right)\right),
\end{equation}
where $\text{Lift}_i$ and $\text{Splat}_i$ project the BEV features onto the projected horizon of camera $i$ (and vice versa) as described above.
Finally, we apply a simple fusion module where we sum the projected features from different cameras together, concatenate them with the lidar features and apply a convolutional block to obtain the final features in BEV.
This simple architecture allows the camera features to be projected from the image plane onto the BEV grid without requiring monocular depth estimation.
We share a single set of transformer weights across all column-frustum pairs and cameras.
For simplicity, we use here a single transformer encoder and decoder but show in~\cref{sec:ablations} that adding more can be beneficial.
All camera features participate in our attention, not just a small number of reference points as is the case in~\cite{li2022bevformer}.
This learnt set of salient features initialises our object detection queries, rather than the fixed maxpool of~\cite{bai2022transfusion}.

\paragraph{Attention vs depth prediction}

It is worth discussing how our approach differs from predicting monocular depth directly. When using monocular depth, each feature in the camera feature map is projected into BEV at multiple locations weighted by a normalised depth distribution. This normalisation limits each feature to be projected either into a single location or smeared with lower intensity across multiple depths. However, in our approach, the attention between camera and lidar is such that the same camera feature can contribute fully to multiple locations in the BEV grid. This is possible because attention is normalised over keys, which correspond to different heights in the camera feature map, rather than queries, which correspond to different distances along the ray. Furthermore, our model has access to lidar features in BEV when choosing where to project camera features, which gives it greater flexibility. Finally, our projection requires fewer parameters than competing methods: 0.9M for our attention-based module compared to 1.6M in the equivalent component of~\cite{liu2022bevfusion}.

\section{Experiments}
We measure the effectiveness of our approach against recent camera-lidar fusion methods that use the Lift-Splat paradigm~\cite{liang2022bevfusion,liu2022bevfusion}.
In all of our experiments, we concentrate on 3D object detection using the nuScenes dataset~\cite{caesar2020nuscenes}, which is a large-scale dataset for autonomous driving.
We use the nuScenes detection score (NDS) and mean average precision (mAP) as evaluation metrics.
We do not consider the extension of~\cite{liang2022bevfusion,liu2022bevfusion} presented in~\cite{ea-bev} as it introduces two supplementary dense depth supervision losses on the camera path to significantly boost the performance of the underlying methods.
In this work, we use solely the 3D object detection losses present in~\cite{bai2022transfusion,liang2022bevfusion,liu2022bevfusion} and leave applying the framework of~\cite{ea-bev} to our method for future work.

\paragraph{Overall architecture}
We use Dual-Swin-Tiny~\cite{Liang_2022} with a feature pyramid network~\cite{lin2017feature} and VoxelNet~\cite{zhou2018voxelnet} as our camera and lidar encoders respectively.
Our object detection head is the transformer-decoder-based module from TransFusion-L~\cite{bai2022transfusion}. 
We use our Lift-Attend-Splat method, described in \cref{sec:method}, to project camera features into BEV space.
We then fuse camera and lidar features using simple concatenation and convolution.
Following~\cite{liu2022bevfusion} the RPN part of VoxelNet is applied to the merged feature.
We ablate alternative choices for the fusion architecture in \cref{sec:ablations}.

\paragraph{Implementation details}
Inputs to the camera encoder have resolution 800x448, which it downsamples by 8x into per-camera feature maps of shape 100x56.
For VoxelNet, we follow the settings of~\cite{liang2022bevfusion}.
We set a maximum of 90k non-empty voxels during training, increased to 180k for inference.
We use an ego-centric BEV grid with dimensions $108\text{m} \times 108\text{m}$ and 0.075m cell size.
This is downsampled 8x by the lidar encoder to the $180\times 180$ grid into which the camera features are projected.
We construct the intermediate projected horizon with 143 uniformly spaced depth bins ranging from 1m to 72m.
For the projection of camera features into BEV, we use the original transformer~\cite{vaswani2017attention} as our encoder-decoder architecture, with one encoder layer, one decoder layer, $d_{\text{model}} = 256$ and $d_{\text{ff}} = 512$.
We replace the ReLU~\cite{nair2010rectified} activation function with GeLU~\cite{hendrycks2016gaussian}, use learnable position embeddings~\cite{gehring2017convolutional} in place of sinusoidal encodings and normalise features before each sublayer~\cite{xiong2020layer}. 
We tie the parameters in each of the 8 heads of our attention blocks.
For the object detection head, we use 200 and 300 queries during training and inference respectively.

\paragraph{Training details}
Our lidar backbone is pretrained on 8 GPUs with batch-size of 1/GPU following the schedule presented in~\cite{bai2022transfusion}, with CBGS~\cite{zhu2019classbalanced} and copy-paste augmentation~\cite{SECOND}.
We initialise the camera backbone with weights pretrained on nuImages~\cite{caesar2020nuscenes} by~\cite{liang2022bevfusion}.
We freeze the lidar backbone and train the camera backbone, projection, fusion and detection head for 20 epochs using 8 GPUs with a batch size of 4/GPU.
We use the AdamW optimiser~\cite{loshchilov2017decoupled} with a maximum learning rate of $5 \times 10^{-5}$ for the camera backbone and $1 \times 10^{-3}$ for all other components. 
We apply the following augmentations: mirroring in the Y dimension, global rotation and scale, and camera-lidar copy-paste~\cite{image-copy-paste}.

\subsection{3D object detection}

We present results for the task of 3D object detection on \cref{tab:detection results}.
Compared to baselines based on the Lift-Splat projection~\cite{liang2022bevfusion,liu2022bevfusion}, our method shows improvements on the validation and test splits of the nuScenes dataset.
In particular, we show substantial improvements in both mAP ($+1.1$) and NDS ($+0.4$) on the test split.
The lidar backbone is frozen and similar in those approaches, showing that our model is better able to leverage camera features.
Our method outperforms the more recent fusion algorithms TransFusion, DeepInteraction, and FUTR3D by a substantial margin, performs similarly to UniTR and MSMDFusion, and slightly underperforms CMT.
These results show that our simple modification of BEVFusion is successful in raising its performance on par with recent SOTA methods.
We leave extending our method with multi-scale feature fusion (used by MSMDFusion) and query denoising (used by CMT) to future work.
Per-class object detection results and comparisons can be found in \cref{tab:full results val} and \cref{sec:suppl:results}.

\begin{figure}
\centering
\includegraphics[width=0.49\linewidth]{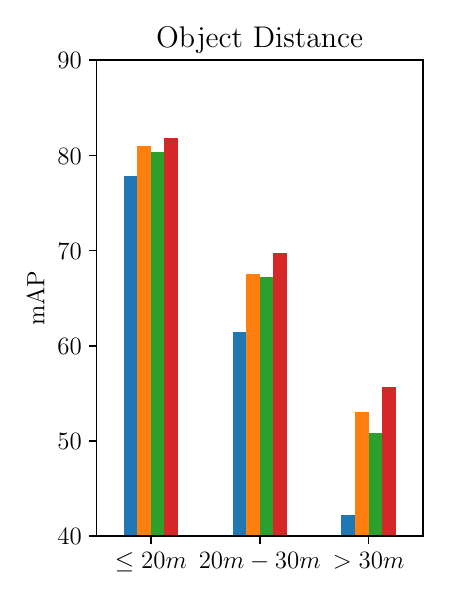}
\includegraphics[width=0.49\linewidth]{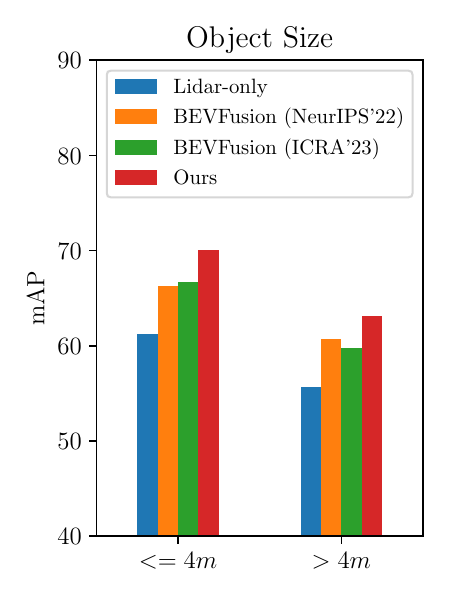}
\caption{\label{fig:buckets}
Object detection performance measured using mAP for objects at different distances from the ego and of different sizes.
Our model consistently outperforms baselines based on Lift-Splat, especially at large distances and for small objects.
}
\end{figure}

We can analyse further the performance of our model by clustering objects together depending on their distances from the ego and on their sizes, see \cref{fig:buckets}.
We see that the bulk of the improvements comes from objects located at large distances and of small sizes.
These are situations for which monocular depth estimation is particularly difficult which could explain why our model fares better in these cases.
Note that even though far-away and small objects contain fewer lidar points, our model is still able to leverage camera features effectively even though the context given by the lidar is weaker.

We show results that use test-time-augmentations (TTA) and model ensembling at the bottom of \cref{tab:detection results}.
We perform TTA over a combination of mirror and rotation augmentations and ensemble models with cell resolutions of 0.05m, 0.075m and 0.10m.
We first apply TTA at each cell resolution and then merge the resulting boxes using Weighted Boxes Fusion (WBF)~\cite{wbf}.
Unsurprisingly, our method shows excellent scaling for these techniques and outperforms BEVFusion~\cite{liu2022bevfusion} on the nuScenes validation set.
More details can be found in \cref{sec:suppl:ensemble}.

\begin{table}
\begin{tabular}{l|cc|cc}
~ & \multicolumn{2}{c|}{val.} & \multicolumn{2}{c}{test} \\
~ & mAP & NDS & mAP & NDS \\
\hline
\hline
BEVFusion~\cite{liu2022bevfusion} & 68.5 & 71.4 & 70.2 & 72.9 \\
BEVFusion~\cite{liang2022bevfusion} & 69.6 & 72.1 & 71.3 & 73.3 \\
FUTR3D~\cite{chen2023futr3d} & 64.2 & 68.0 & 69.4 & 72.1 \\
TransFusion~\cite{bai2022transfusion} & 67.5 & 71.3 & 68.9 & 71.6 \\
DeepInteraction~\cite{yang2022deepinteraction} & 69.9 & 72.6 & 70.8 & 73.4  \\
MSMDFusion~\cite{jiao2023msmdfusion} & - & - & 71.5 & 74.0  \\
CMT~\cite{yan2023cross} & 70.3 & 72.9 & 72.0 & 74.1  \\
UniTR~\cite{wang2023unitr} & 70.5 & 73.3 & 70.9 & 74.5 \\
\hline
Ours                   & 71.2 & 72.7 & 71.5 & 73.6 \\
Ours w/ TFA            & 72.1 & 73.8 & - & - \\
\hline
\hline
BEVFusion$^\ddagger$~\cite{liu2022bevfusion} & 73.7 & 74.9 & 75.0 & 76.1 \\
\hline
Ours$^\ddagger$        & 74.6 & 75.1 & - & - \\
Ours w/ TFA$^\ddagger$ & 75.7 & 76.0 & 75.5 & 74.9 \\
\end{tabular}
\caption{\label{tab:detection results}
Object detection performance on the validation and test splits of the nuScenes dataset.
TFA: Temporal Feature Aggregation. $^\ddagger$ denotes test-time augmentations and model ensembling.
}
\end{table}

\subsection{Qualitative analysis} \label{sec:quality}

\begin{figure*}[ht!]
\begin{center}
\begin{minipage}{0.61\textwidth}
  \begin{subfigure}{0.49\textwidth}
    \centering
    \caption*{Ours}
    \begin{tikzpicture}
        \node[anchor=south west, inner sep=0] (image) at (0,0) {
            \includegraphics[width=\linewidth]{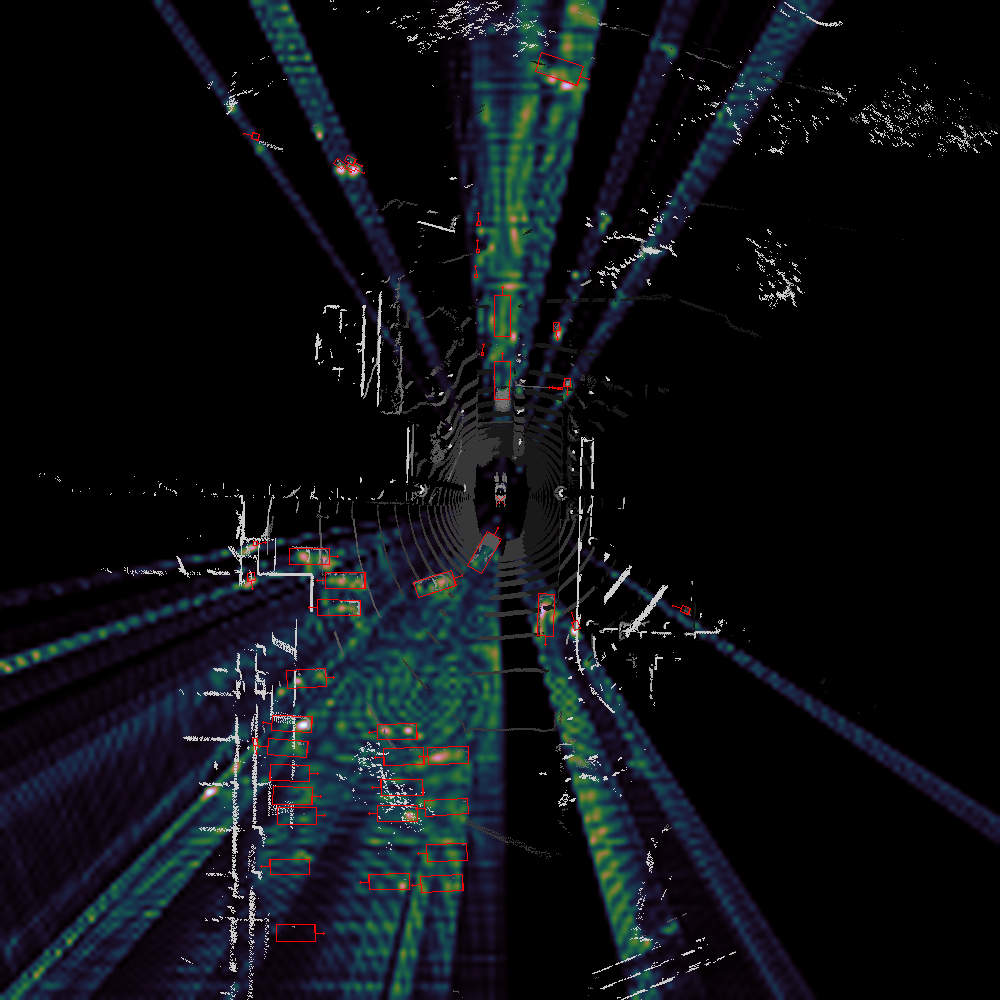}
        };
        \begin{scope}[x={(image.south east)},y={(image.north west)}]
          \draw[white,thick,densely dotted] (0.275, 0.365) rectangle ++(.1, .1);
          \draw[orange,thick,densely dotted] (0.3, 0.79) rectangle ++(.1, .1);
        \end{scope}
    \end{tikzpicture}
  \end{subfigure}
  \begin{subfigure}{0.49\textwidth}
    \centering
    \caption*{BEVFusion~\cite{liu2022bevfusion}}
    \begin{tikzpicture}
        \node[anchor=south west, inner sep=0] (image) at (0,0) {
            \includegraphics[width=\linewidth]{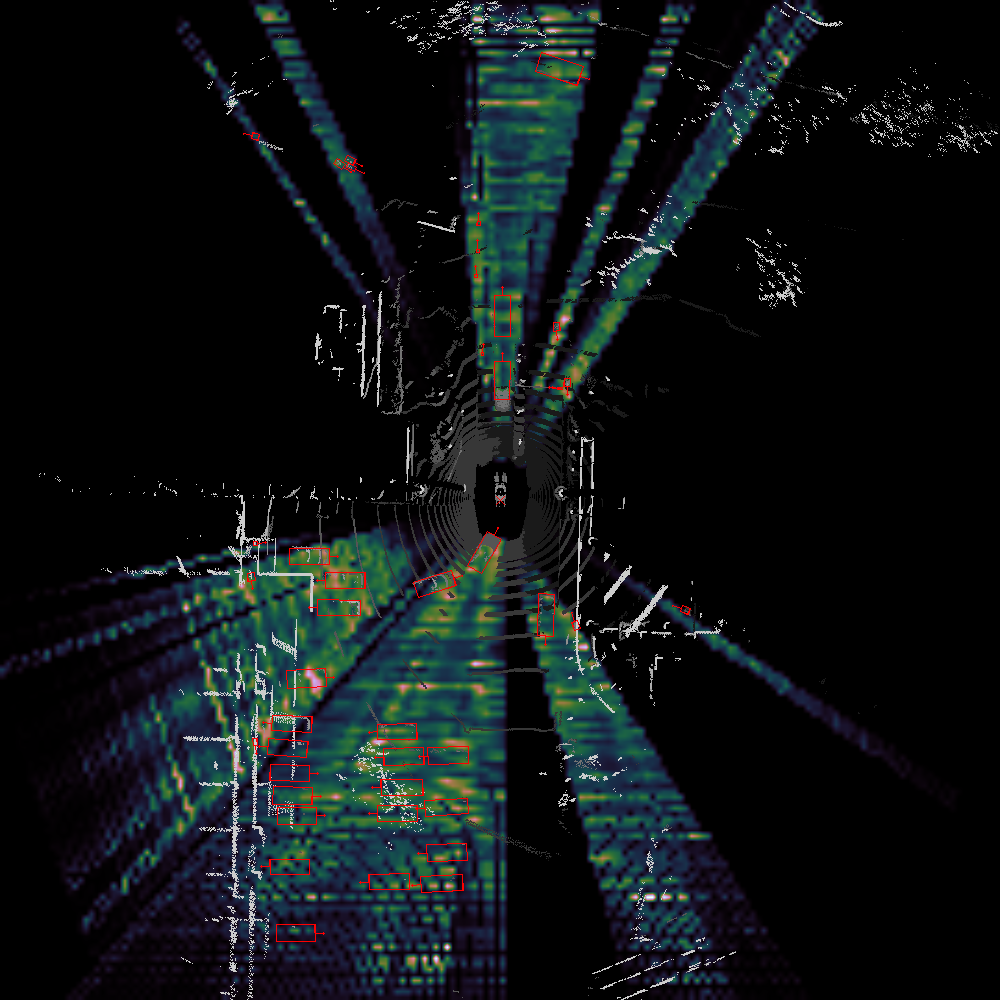}
        };
        \begin{scope}[x={(image.south east)},y={(image.north west)}]
          \draw[white,thick,densely dotted] (0.275, 0.365) rectangle ++(.1, .1);
          \draw[orange,thick,densely dotted] (0.3, 0.79) rectangle ++(.1, .1);
        \end{scope}
    \end{tikzpicture}
  \end{subfigure}
  \vspace{0.15em}
  \\
  \begin{subfigure}{0.49\textwidth}
    \centering
    \begin{tikzpicture}
        \node[anchor=south west, inner sep=0] (image) at (0,0) {
            \includegraphics[trim={275px 365px 625px 535px}, clip, width=0.49\linewidth]{images/explainability_final/bev_attention_bs4_e19_scene-0003_step8.png}
        };
        \begin{scope}[x={(image.south east)},y={(image.north west)}]
          \draw[white,very thick,densely dotted] (0.01, 0.01) rectangle (0.99, 0.99);
        \end{scope}
    \end{tikzpicture}
    \hfill
    \begin{tikzpicture}
        \node[anchor=south west, inner sep=0] (image) at (0,0) {
            \includegraphics[trim={300px 790px 600px 110px}, clip, width=0.49\linewidth]{images/explainability_final/bev_attention_bs4_e19_scene-0003_step8.png}
        };
        \begin{scope}[x={(image.south east)},y={(image.north west)}]
          \draw[orange,very thick,densely dotted] (0.01, 0.01) rectangle (0.99, 0.99);
        \end{scope}
    \end{tikzpicture}
    \caption{\label{sec:explainability:bevlas}}
  \end{subfigure}
  \begin{subfigure}{0.49\textwidth}
    \centering
    \begin{tikzpicture}
        \node[anchor=south west, inner sep=0] (image) at (0,0) {
            \includegraphics[trim={275px 365px 625px 535px}, clip, width=0.49\linewidth]{images/explainability_final/bev_attention_mit_scene-0003_step8.png}
        };
        \begin{scope}[x={(image.south east)},y={(image.north west)}]
          \draw[white,very thick,densely dotted] (0.01, 0.01) rectangle (0.99, 0.99);
        \end{scope}
    \end{tikzpicture}
    \hfill
    \begin{tikzpicture}
        \node[anchor=south west, inner sep=0] (image) at (0,0) {
            \includegraphics[trim={300px 790px 600px 110px}, clip, width=0.49\linewidth]{images/explainability_final/bev_attention_mit_scene-0003_step8.png}
        };
        \begin{scope}[x={(image.south east)},y={(image.north west)}]
          \draw[orange,very thick,densely dotted] (0.01, 0.01) rectangle (0.99, 0.99);
        \end{scope}
    \end{tikzpicture}
    \caption{\label{sec:explainability:bevlss}}
  \end{subfigure}
\end{minipage}
\hfill
\begin{minipage}{0.36\textwidth}
  \vspace{-1.1em}
  \begin{subfigure}{\textwidth}
    \centering
    \caption*{Camera Image}
    \includegraphics[trim={200px 0 160px 160px}, clip, width=0.49\linewidth]{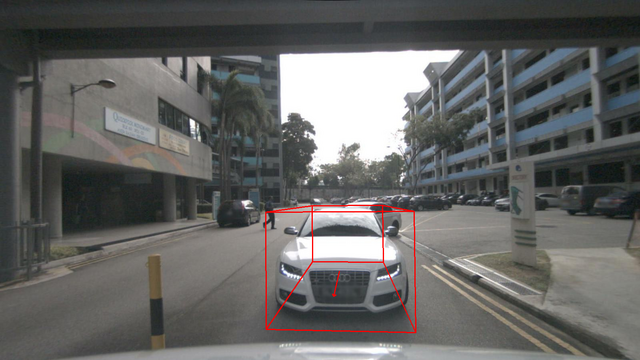}
  \end{subfigure}
  \vspace{0.1em}\\
  \begin{subfigure}{0.49\textwidth}
    \centering
    \caption*{Ours}
    \includegraphics[trim={200px 0 160px 160px}, clip, width=\linewidth]{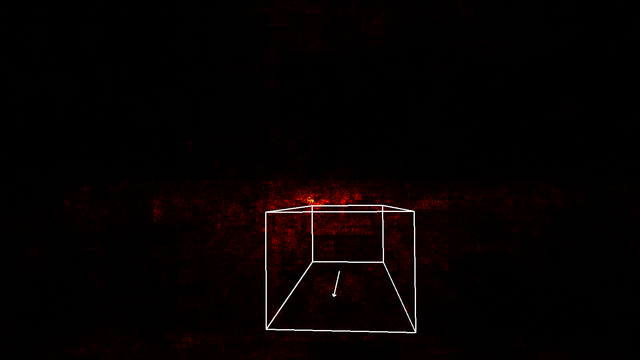}
  \end{subfigure}
  \begin{subfigure}{0.49\textwidth}
    \centering
    \caption*{BEVFusion~\cite{liu2022bevfusion}}
    \includegraphics[trim={200px 0 160px 160px}, clip, width=\linewidth]{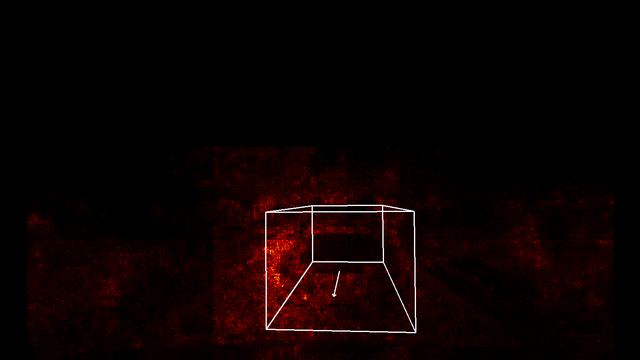}
  \end{subfigure}
  \\
  \begin{subfigure}{0.49\textwidth}
    \centering
    \includegraphics[trim={200px 0 160px 160px}, clip, width=\linewidth]{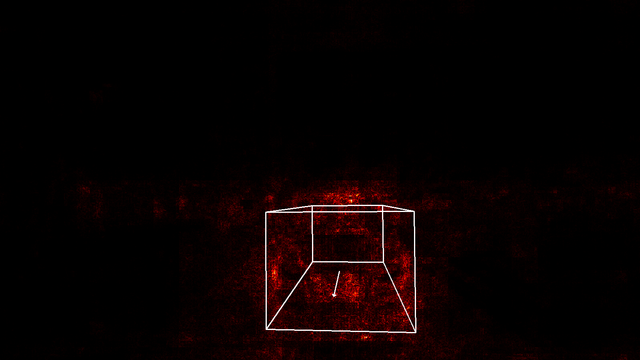}
  \end{subfigure}
  \begin{subfigure}{0.49\textwidth}
    \includegraphics[trim={200px 0 160px 160px}, clip, width=\linewidth]{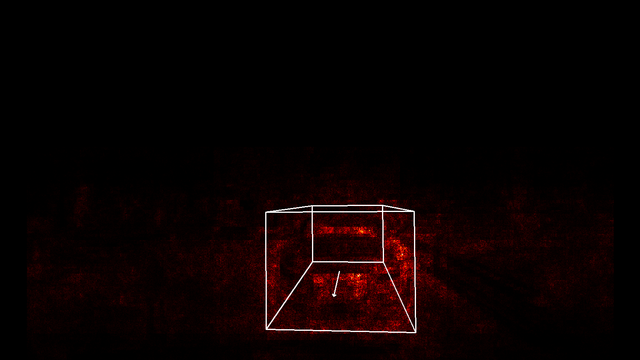}
  \end{subfigure}
  \\
  \begin{subfigure}{\textwidth}
  \caption{\label{fig:explainability:saliency}}
  \end{subfigure}
\end{minipage}
\begin{minipage}{0.02\textwidth}
  \begin{subfigure}{\textwidth}
    \caption*{\rotatebox{90}{Camera+lidar ~~ \qquad \qquad \qquad \qquad}} 
  \end{subfigure}
  \\
  \begin{subfigure}{\textwidth}
    \caption*{\rotatebox{90}{Camera only ~~~ \quad}} 
  \end{subfigure}
\end{minipage}
\end{center}
\caption{\label{fig:explainability}
\textbf{(a, b)}
Visualisation of where camera features of ground-truth objects are projected onto the BEV grid for our method compared to BEVFusion~\cite{liu2022bevfusion}.
We observe that our method is able to place camera features around objects more narrowly than BEVFusion, which is based on monocular depth estimation.
\textbf{(c)}
Comparison of saliency maps, cropped to aid visualisation, given the camera image (top) for models trained with camera-lidar (middle) and camera only (bottom).
When trained with both camera and lidar, our model selects camera features in an area that is different than when trained with camera only, while~\cite{liu2022bevfusion} behaves similarly in both settings.
} 
\end{figure*}

We visualise where camera features are projected onto the BEV grid and compare our method to BEVFusion~\cite{liu2022bevfusion}.
For our method, we examine the attention map of the final cross-attention block in the transformer, averaged over all attention heads.
For BEVFusion, we use the the monocular depth estimate to establish the strength of correspondence between positions in camera and BEV space. 
We consider only the pixels corresponding to ground-truth objects when calculating the total weight of projected camera features in BEV, see \cref{sec:suppl:attention} for details.
As can be seen in \cref{sec:explainability:bevlas}, our method places camera features predominantly in regions where ground-truth bounding boxes are present.
This indicates that it can effectively leverage the lidar point cloud as a context to project camera features at the relevant location in BEV.
Compared to BEVFusion shown in \cref{sec:explainability:bevlss}, the distribution of features appears more narrowly localised and stronger around objects.
This could be because our projection mechanism does not require the weights of the camera features to be normalised along their ray, giving our model more flexibility to place features at the desired location.
Note that, even though our method also projects camera features outside of ground-truth boxes in BEV, the strength of the activation in these regions is suppressed by the fusion module, see \cref{fig:appendix:bev activations}.
This is consistent with our findings in \cref{sec:depth}, where we showed that the latter part of the model can effectively suppress camera features at the wrong location.
More examples can be found in \cref{sec:suppl:attention}.

We further explore which pixels in the camera images are most attended to using saliency maps~\cite{simonyan2013deep}.
These are derived by computing the gradient of the maximum class logit with respect to a camera image $I_j$, given object query index $i$ and probabilities $z$, as $\partial z_{i, \hat c} / \partial I \vert_{I_j}$ where $\hat c=\arg \max_{c} z_{i, c}$.
They allow us to visualise the contribution of individual pixels to the final detection for a selected object, see \cref{fig:explainability:saliency}.
Interestingly, we observe that when trained with both camera and lidar, our model tends to select camera features at different locations than when trained with cameras only.
In the absence of lidar, our method selects camera features across the entirety of the object, while in the presence of both lidar and cameras, the model selects camera features mainly from the upper part of the object.
We observe that this pattern is mostly prevalent for objects close to the ego which are well-represented by lidar point clouds but fades away for far-away objects or objects with few lidar points such as pedestrians, see \cref{fig:appendix:explainability} for more examples.
We hypothesise that our projection architecture enables the model to select camera features that best complement the information encoded in lidar, resulting in differing attention patterns between camera-only and fusion settings.
This pattern is less present in BEVFusion~\cite{liu2022bevfusion}, which attends to the broader neighbourhood of pixels surrounding the selected object in both cases. 

\subsection{Temporal feature aggregation} \label{sec:aggregation}
Because our method fuses camera and lidar features onto a BEV grid, we can easily leverage past information using temporal feature aggregation (TFA).
To achieve this, we implement the simple autoregressive procedure of VideoBEV~\cite{han2023exploring} but aggregate the fused BEV features $B^{\text{fus.}}$ instead of the camera features.
Our method is as follows:
(i) save the fused BEV features from the previous timestep,
(ii) apply ego motion compensation to align these features with the current timestep, using bilinear sampling to construct the new feature grid,
(iii) concatenate these features with the fused BEV features of the current timestep and merge them using a simple $3\times 3$ convolutional block.

We train TFA models on sequences of 3 frames for 10 epochs starting from our single frame model's object detection head, lidar and camera backbones.
During training, the lidar and camera backbones are frozen.
For inference, we accumulate BEV features for the entire length of a run, yielding detections at each time step.
\Cref{tab:detection results} shows that temporal feature aggregation boosts object detection performance significantly in all configurations.

\subsection{Ablation experiments} \label{sec:ablations}
\begin{table}
\centering
\begin{tabular}{llcc}
~ & ~ & mAP & NDS   \\
\hline\hline
\textbf{Fusion module}  & Cat+Conv                                  & \textbf{70.43} & 71.9 \\
                        & Gated sigmoid~\cite{liang2022bevfusion}   & 70.12          & 71.9 \\
                        & Add                                       & 70.32          & \textbf{72.1} \\
\hline
\textbf{\# decoder blocks$^*$}  & 1 block                                   & 70.29          & 71.9 \\
                                & 2 blocks                                  & 70.40          & \textbf{72.0} \\
                                & 4 blocks                                  & \textbf{70.49} & 71.9 \\
\hline
\textbf{\# TFA frames}  & 1 frame (no TFA)                           & 71.2           & 72.8        \\
                        & 2 frames                                  & 72.1           & 73.3        \\
                        & 3 frames                                  & \textbf{72.1}  & \textbf{73.8}
\caption{\label{tab:ablation}
Impact of model modifications on 3D object detection performance: (i) feature fusion module, (ii) number of transformer decoder blocks in the ``Attend'' stage, (iii) number of frames in Temporal Feature Aggregation (TFA).
$^*$~frozen camera backbone.
}
\end{tabular}
\end{table}

We ablate some design choices for our method and show their impact on object detection performance on \cref{tab:ablation}. 
All ablation experiments use a simpler training schedule with 10 epochs, batch accumulation instead of full batch training and no camera augmentations.
We first analyse the effect of different fusion modules: we compare a simple skip connection (add), a small concatenation and convolution layer (Cat+Conv as in~\cite{liu2022bevfusion}) and a gated sigmoid block~\cite{liang2022bevfusion}. We find all perform similarly, with Cat+Conv achieving the best mAP, contrary to findings of~\cite{liang2022bevfusion}.
We also ablate the number of transformer decoder blocks in the ``Attend'' stage of our projection and show that increasing their number leads to a small improvement in mAP.
This suggests that our method's performance scales with compute.
We use a single decoder block in our main experiments to balance quality and performance.
Finally, we see good improvement in NDS with an increasing number of frames in TFA during training.
\section{Conclusion}
In this work, we analysed the role of monocular depth prediction in recent state-of-the-art camera-lidar fusion methods 
and showed that, surprisingly, improvements in depth estimation did not lead to better object detection performance.
Strikingly, we also showed that removing depth estimation altogether did not worsen performance significantly.
This led us to hypothesise that relying on monocular depth estimation could be an unnecessary architectural bottleneck when fusing camera and lidar,
and prompted us to introduce a novel fusion method that directly combines camera and lidar features using a simple attention mechanism.
Compared to projecting camera features using monocular depth, our method allows camera features to contribute to multiple locations in BEV space and gives our model greater flexibility to select complementary camera and lidar features.
Finally, we validated the effectiveness of our method on the nuScenes dataset and showed that it improves object detection performance over baselines based on monocular depth estimation and showcased the role of attention as a key contributor to these improvements.
We leave detailed investigation of our model in the camera-only setting and inclusion of radar as future work.
We hope that our findings will motivate discussions around the role of monocular depth prediction in camera-lidar fusion and prompt further developments in multimodal perception.
\section*{Acknowledgements}
We thank our FiveAI and Bosch colleagues, especially Tom Joy and Anthony Knittel, for their valuable feedback.
We are grateful to Bhavesh Garg for his initial experiments into depth supervision and to Blaine Rogers for initiating work on multimodal fusion at FiveAI.
\FloatBarrier
{
    \small
    \bibliographystyle{ieeenat_fullname}
    \bibliography{main}
}

\FloatBarrier
\clearpage
\clearpage
\maketitlesupplementary

\setcounter{section}{0}
\setcounter{table}{0}
\setcounter{figure}{0}

\renewcommand{\thesection}{\Alph{section}}
\renewcommand{\thetable}{S\arabic{table}}
\renewcommand{\thefigure}{S\arabic{figure}}

\section{Monocular depth in the ``LiftSplat'' paradigm}

\subsection{Computation of ground truth depth from lidar}  \label{sec:suppl:gtdepth}

For each camera image we compute the ground depth map $D^{\text{gt}}\in \mathbb{R}^{H \times W}$ by projecting the 3D lidar point cloud onto the image plane and binning each point within the pixels of the camera feature map.
For non-empty cells, we follow~\cite{li2022bevdepth} and choose the depth to be the minimum distance (from the camera plane) of all the points in the cell, leaving the depth unspecified for empty cells and those for which the minimum yields a depth value which is outside the range of the model's depth bins.
This depth map is suitable for visualisation and depth metric evaluation, but for depth supervision it is necessary to calculate the one-hot encoding of $D^{\text{gt}}$ according to buckets defined by the model's depth bins $d \in \mathbb{R}^{N_D}$.

\subsection{Visualisation of depth maps} \label{sec:suppl:depth viz}

We generate the monocular depth maps shown in~\cref{fig:depth quality} by calculating the weighted average of the model's depth bins $d$ with the predicted depth distribution $D^{\text{pred}} \in \mathbb{R}^{N_D \times H \times W}$
\begin{equation}
D^{\text{mean}}_{h, w} = \sum^{N_D}_{n}{d_n D^{\text{pred}}_{n,h,w}}.
\end{equation}
This depth map is constrained by construction to $[\text{min}(d), \text{max}(d)]$ and we map this range onto the Turbo colour map~\cite{mikhailov2019turbo} for visualisation.

The lidar depth map $D^{\text{gt}}$ is similarly colourised, except for cells where the depth is unspecified as described above (see ~\cref{fig:superviseddepthloss}, top-right) which are coloured grey.

\subsection{Supervision of predicted depth using lidar} \label{sec:suppl:depth supervision}

We perform all of our experiments using the method presented in~\cite{liu2022bevfusion} and use the original repository\footnote{\url{https://github.com/mit-han-lab/bevfusion}}.
We use the vanilla Lift-Splat transform implemented in the class \texttt{LSSTransform} with default parameters provided in the original work.
We supervise the depth classifier by introducing the following loss alongside the original detection losses,
\begin{equation}
\label{eq:depth loss}
    L_{\text{depth}} = -\frac{1}{N}\sum_n^N\text{log} (D_{n} \cdot 1_n),
\end{equation}
which is a cross-entropy loss between the lidar depth distribution and predicted depth distribution, taken over all cells for which the lidar depth is available. $D_n \in \mathbb{R}^{N_D}$ is the normalised predicted depth distribution from the LiftSplat model for the $n$th cell, $1_n$ is the one-hot encoded lidar depth distribution for the $n$th cell.
The model is trained end-to-end with all components unfrozen as in~\cite{liu2022bevfusion} and hyper-parameter $\lambda$ controlling the strength of the depth supervision loss with respect to the detection losses.

We also experiment with pretraining the depth estimation module within LiftSplat.
First, we train the camera stream in~\cite{liu2022bevfusion} supervising only the depth distribution with the whole camera pipeline unfrozen.
Following this pretraining, we add the lidar components and train the full model end-to-end as in~\cite{liu2022bevfusion}, with no depth supervision ($\lambda = 0$) and all modules unfrozen.
\begin{table*}
\begin{center}
\begin{tabular}{|c|c||c|c|c|c|c|c|c|c|c|c|}
\hline
\multirow{3}{*}{Loss Weight} & \multirow{3}{*}{mAP} & \multicolumn{5}{c}{mode} & \multicolumn{5}{|c|}{mean} \\\cline{3-7}\cline{8-12}
&& \multicolumn{2}{c}{Relative} & \multicolumn{2}{|c|}{RMSE} & \multirow{2}{*}{Frac. 125} & \multicolumn{2}{c}{Relative} & \multicolumn{2}{|c|}{RMSE} & \multirow{2}{*}{Frac. 125} \\\cline{3-6}\cline{8-11}
&& Abs. & Sq. & Linear & Log && Abs. & Sq. & Linear & Log & \\
\hline\hline
BEVFusion~\cite{liu2022bevfusion} & \textbf{68.5} & 2.95 & 133.76 & 25.95 & 1.87 & 0.97 & 2.75 & 61.31 & 17.40 & 1.30 & 0.88 \\
\hline
$\lambda = 0$ & 68.4 & 3.69 & 176.09 & 30.22 & 1.90 & 0.96 & 2.83 & 68.73 & 18.54 & 1.34 & 0.87 \\
0.001 & 68.1 & 1.79 & 65.63 & 20.16 & 1.77 & 0.94 & 3.14 & 79.87 & 19.91 & 1.39 & 0.88 \\
0.01 & 68.0 & 0.61 & 11.78 & 11.54 & 1.03 & 0.63 & 0.76 & 10.30 & 8.09 & 0.68 & 0.61 \\
0.1 & 68.1 & 0.38 & 5.53 & 9.28 & 0.77 & 0.41 & 0.43 & 4.97 & 6.47 & 0.46 & 0.37 \\
1 & 68.1 & 0.21 & 2.48 & 5.78 & 0.37	& 0.20 & 0.22 & 2.23 & 4.77 & 0.33 & 0.19 \\
5 & 66.6 & 0.19 & 2.01 & 4.77 & 0.33 & 0.17 & 0.19 & 1.95 & 4.53 & 0.32 & 0.17 \\
100 & 64.6 & 0.16 & 1.15 & 4.64 & 0.33 & 0.17 & 0.16 & 1.12 & 4.55 & 0.32 & 0.17 \\
\hline
Pretrained & 67.4 & 0.54 & 8.10 & 9.95 & 0.86 & 0.61 & 0.64 & 7.91 & 7.87 & 0.66 & 0.57 \\
Lidar & 68.4 & 0.04 & 0.01 & 0.29 & 0.05 & 0.00 & 0.04 & 0.01 & 0.29 & 0.05 & 0.00 \\
Uniform depth & \textbf{68.5} & -- & -- & -- & -- & -- & -- & -- & -- & -- & -- \\
\hline
\end{tabular}
\end{center}
\caption{\label{tab:superviseddepthloss}
Extended analysis of the monocular depth quality provided by different variations of the ``LiftSplat'' camera feature projection, see \cref{sec:suppl:depth results}.
}
\end{table*}

\subsection{Extended depth quality results} \label{sec:suppl:depth results}

\begin{figure*}
\begin{center}
  \begin{subfigure}{0.48\textwidth}
    \centering
    \includegraphics[width=\linewidth]{images/depth_cam_crop_reduced.png}
    \caption*{Camera}
  \end{subfigure}
  \begin{subfigure}{0.48\textwidth}
    \centering
    \includegraphics[width=\linewidth]{images/depth_lidar.png}
    \caption*{Lidar}
  \end{subfigure}
  \\
  \begin{subfigure}{0.48\textwidth}
    \centering
    \includegraphics[width=\linewidth]{images/depth_e2e.png}
    \caption*{End-to-end~\cite{liu2022bevfusion}}
  \end{subfigure}
  \begin{subfigure}{0.48\textwidth}
    \centering
    \includegraphics[width=\linewidth]{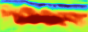}
    \caption*{Supervised $\lambda = 0$}
  \end{subfigure}
  \\
  \begin{subfigure}{0.48\textwidth}
    \centering
    \includegraphics[width=\linewidth]{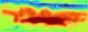}
    \caption*{Supervised $\lambda = 0.001$}
  \end{subfigure}
  \begin{subfigure}{0.48\textwidth}
    \centering
    \includegraphics[width=\linewidth]{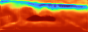}
    \caption*{Supervised $\lambda = 0.01$}
  \end{subfigure}
  \\
  \begin{subfigure}{0.48\textwidth}
    \centering
    \includegraphics[width=\linewidth]{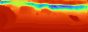}
    \caption*{Supervised $\lambda = 0.1$}
  \end{subfigure}
  \begin{subfigure}{0.48\textwidth}
    \centering
    \includegraphics[width=\linewidth]{images/depth_sup_1.png}
    \caption*{Supervised $\lambda = 1$}
  \end{subfigure}
  \\
  \begin{subfigure}{0.48\textwidth}
    \centering
    \includegraphics[width=\linewidth]{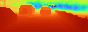}
    \caption*{Supervised $\lambda = 5$}
  \end{subfigure}
  \begin{subfigure}{0.48\textwidth}
    \centering
    \includegraphics[width=\linewidth]{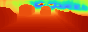}
    \caption*{Supervised $\lambda = 100$}
  \end{subfigure}
\end{center}
\caption{Depth maps obtained after different levels of depth supervision on an example from the nuScenes val set.}
\label{fig:superviseddepthloss}
\end{figure*}

We evaluate the performance of the depth classifier using five of the metrics proposed in~\cite{eigen2014depth}: root mean squared error (RMSE), root mean squared logarithmic error (RMSLE), mean absolute relative error (Abs. Rel.), mean squared relative error (Sq. Rel.) and fraction outside 125\% (Frac. 125).
We show all the metrics for 2 different methods of translating the classification output into a depth map: ``mode'' --- in which we use the bin with maximum probability, and ``mean'' --- where the predicted depth is the weighted average of all the bins.
We take the average of these quantities over all predictions made by the depth classifier for which we have ground truth to compute the metrics.
Camera feature cells for which lidar depth is unspecified are ignored.
We compare BEVFusion and four different variants: adding depth supervision using \cref{eq:depth sup loss} with various weights $\lambda$, using lidar depth maps instead of monocular depth estimation (lidar), using a pretrained and frozen depth classifier (pretrained), and finally removing depth estimation altogether by projecting camera features at all depths uniformly using \cref{eq:swiftblat} (uniform depth).
Quantitative results can be seen in~\cref{tab:superviseddepthloss} and qualitative comparisons in~\cref{fig:superviseddepthloss}.

\section{Detailed experimental results}

\subsection{3D object detection} \label{sec:suppl:results}
\begin{table*}
\centering
\begin{tabular}{|l|c|c|c|c|c|c|c|c|c|c||c|c|}
        \hline
        Model & barrier & bicycle & bus & car & CV & MC & ped & TC & trailer & truck & mAP & NDS\\
        \hline \hline
        Ours & 74.1 & 70.0 & 81.3 & 90.3 & 33.8 & 80.8 & 89.3 & 79.7 & 44.0 & 68.2 & 71.2 & 72.7\\
        BEVFusion\cite{liang2022bevfusion} & 73.5 & 67.5 & 77.7 & 89.1 & 30.9 & 79.0 & 89.4 & 79.3 & 42.6 & 66.7 & 69.6 & 72.1\\
        DeepInteraction\cite{yang2022deepinteraction} & 78.1 & 52.9 & 68.3 & 87.1  & 33.1 & 73.6 & 88.4 & 86.7 & 60.8 & 60.0 & 69.9 & 72.6\\
        Ours$^\ddagger$ & 77.5 & 75.2 & 82.3 & 91.2 & 40.0 & 85.6 & 90.6 & 80.2 & 50.1 & 72.2 & 74.6 & 75.1 \\
        Ours w/ TFA & 74.4 & 72.4 & 81.6 & 90.8 & 33.7 & 82.5 & 89.8 & 79.6 & 45.8 & 70.1 & 72.1 & 73.8 \\
        Ours$^\ddagger$ w/ TFA & 78.6 & 78.2 & 84.3 & 91.6 & 39.9 & 87.5 & 91.4 & 80.7 & 51.2 & 73.3 & 75.7 & 76.0 \\
        \hline \hline
        Ours & 78.0 & 54.9 & 72.1 & 89.0 & 38.9 & 75.3 & 90.3 & 87.0 & 65.3 & 64.2 & 71.5 & 73.6 \\
        Ours$^\ddagger$ w/ TFA & 79.7 & 65.2 & 75.2 & 90.3 & 43.5 & 82.8 & 92.0 & 87.1 & 70.1 & 68.9 & 75.5 & 74.9 \\
        BEVFusion\cite{liang2022bevfusion} & 78.3 & 56.5  & 72.0 & 88.5 & 38.1 & 75.2 & 90.0 & 86.5 & 64.7 & 63.1 & 71.3 & 73.3 \\
        DeepInteraction\cite{yang2022deepinteraction} & 80.4 & 54.5 & 70.8 & 87.9 & 37.5 & 75.4 & 91.7 & 87.2 & 63.8  & 60.2 & 70.8 & 73.4 \\
        \hline
    \end{tabular}
\caption{Per-class object detection scores on the nuScenes validation set (top) and test set (bottom). TFA: Temporal Feature Aggregation. $^\ddagger$ indicates ensembling + TTA.}
\label{tab:full results val}
\end{table*}
In ~\cref{tab:full results val} we present per-class detection scores and compare our model to other state-of-the-art models on the validation and test splits of the nuScenes dataset. Our method outperforms baselines based on the LiftSplat paradigm.
We are additionally showing how test-time-augmentations and temporal feature aggregation further improves these results.

\subsection{Detailed qualitative results} \label{sec:suppl:attention}

\begin{figure*}
\begin{center}
\begin{minipage}{0.48\textwidth}
  \begin{subfigure}{0.49\textwidth}
    \centering
    \caption*{Ours}
    \includegraphics[trim={225px 0 225px 200px}, clip, width=\linewidth]{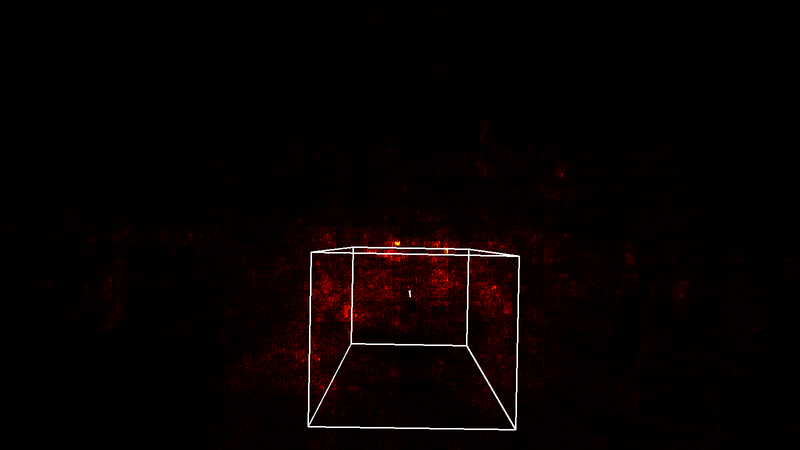}
  \end{subfigure}
  \begin{subfigure}{0.49\textwidth}
    \centering
    \caption*{BEVFusion~\cite{liu2022bevfusion}}
    \includegraphics[trim={225px 0 225px 200px}, clip, width=\linewidth]{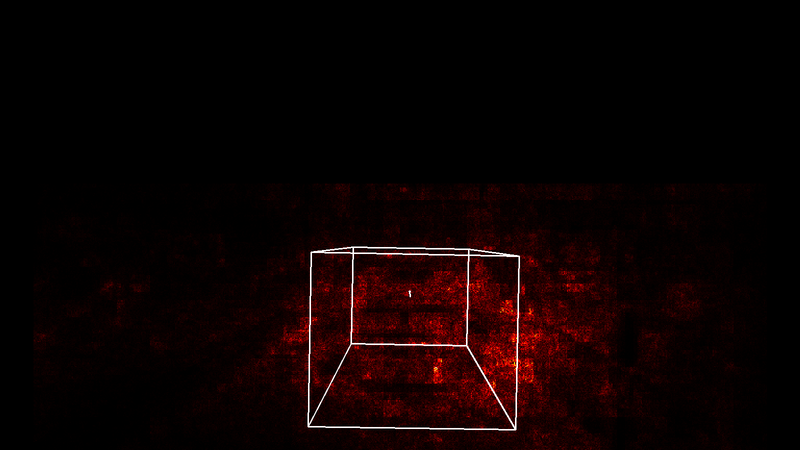}
  \end{subfigure}
  \\
  \begin{subfigure}{0.49\textwidth}
    \centering
    \includegraphics[trim={225px 0 225px 200px}, clip, width=\linewidth]{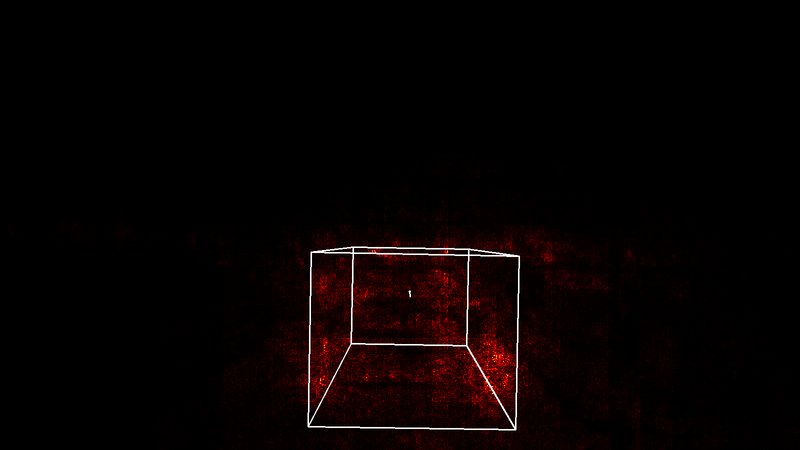}
  \end{subfigure}
  \begin{subfigure}{0.49\textwidth}
    \includegraphics[trim={225px 0 225px 200px}, clip, width=\linewidth]{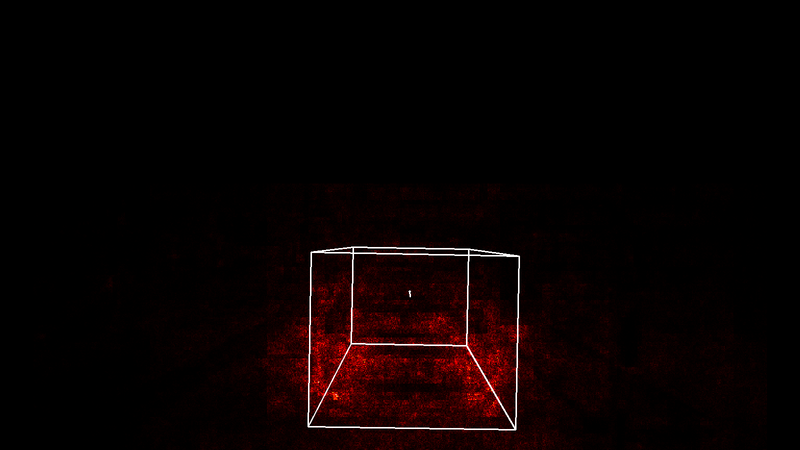}
  \end{subfigure}
  \\
  \begin{subfigure}{\textwidth}
  \centering
  \begin{tikzpicture}
    \node[anchor=south west,inner sep=0] at (0,0) {
    \includegraphics[width=0.6\linewidth]{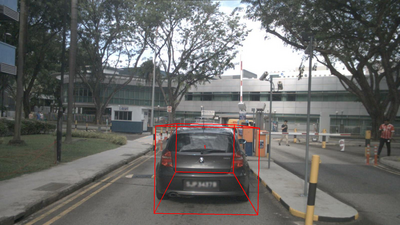}
    };
    \draw[white,dashed,thick] (0.1687\linewidth,0.003\linewidth) rectangle (0.431\linewidth, 0.1875\linewidth);
    \end{tikzpicture}
    \caption{}
  \end{subfigure}
\end{minipage}
\begin{minipage}{0.48\textwidth}
  \begin{subfigure}{0.49\textwidth}
    \centering
    \caption*{Ours}
    \includegraphics[trim={325px 107px 275px 200px}, clip, width=\linewidth]{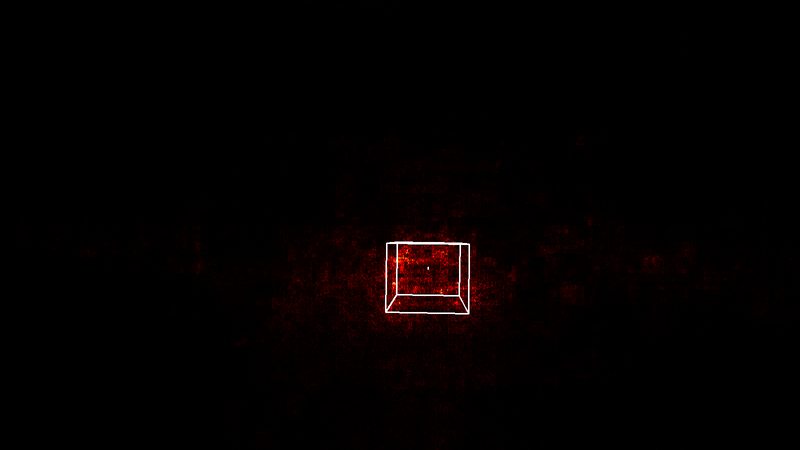}
  \end{subfigure}
  \begin{subfigure}{0.49\textwidth}
    \centering
    \caption*{BEVFusion~\cite{liu2022bevfusion}}
    \includegraphics[trim={325px 107px 275px 200px}, clip, width=\linewidth]{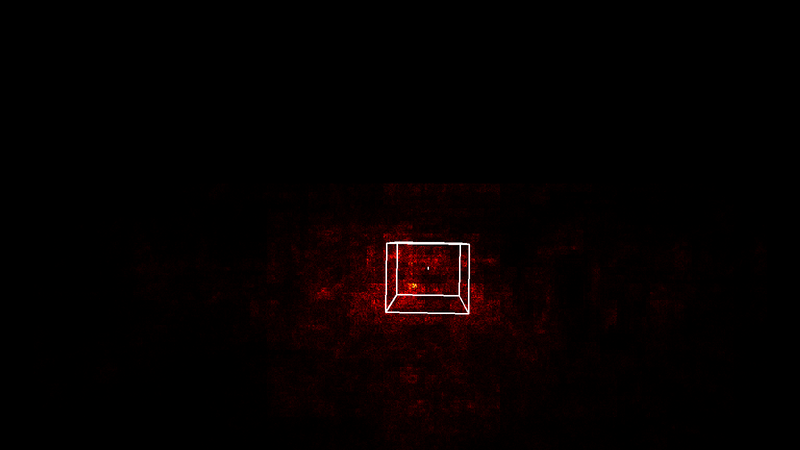}
  \end{subfigure}
  \\
  \begin{subfigure}{0.49\textwidth}
    \centering
    \includegraphics[trim={325px 107px 275px 200px}, clip, width=\linewidth]{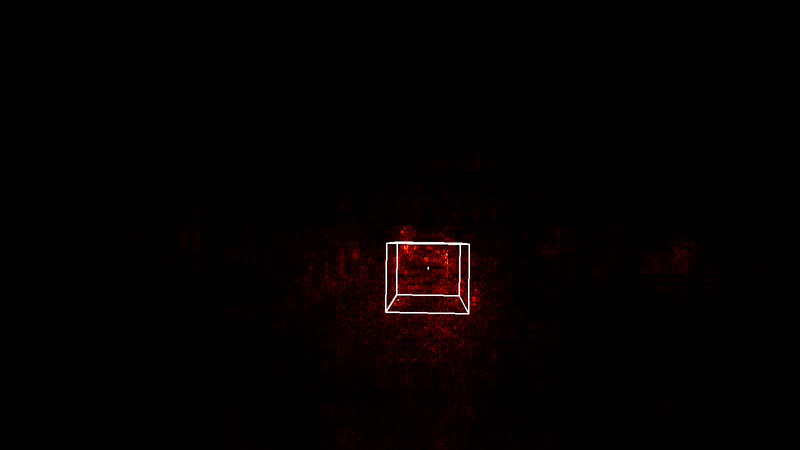}
  \end{subfigure}
  \begin{subfigure}{0.49\textwidth}
    \includegraphics[trim={325px 107px 275px 200px}, clip, width=\linewidth]{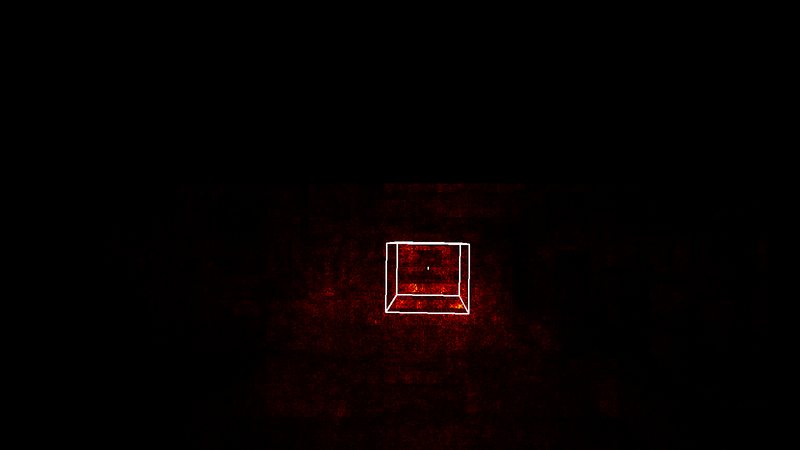}
  \end{subfigure}
  \\
  \begin{subfigure}{\textwidth}
  \centering
  \begin{tikzpicture}
    \node[anchor=south west,inner sep=0] at (0,0) {
    \includegraphics[width=0.6\linewidth]{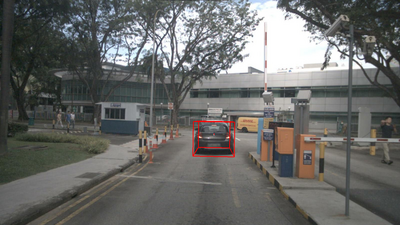}
    };
    \draw[white,dashed,thick] (0.24375\linewidth,0.08\linewidth) rectangle (0.39375\linewidth, 0.1875\linewidth);
    \end{tikzpicture}
    \caption{}
  \end{subfigure}
\end{minipage}
\begin{minipage}{0.02\textwidth}
  \centering
  \begin{subfigure}{\textwidth}
    \caption*{\rotatebox{90}{\qquad \qquad \quad Cam+lidar}} 
  \end{subfigure}
  \\
  \begin{subfigure}{\textwidth}
    \caption*{\rotatebox{90}{\qquad \qquad \qquad \qquad \qquad Cam-only}} 
  \end{subfigure}
\end{minipage}
\begin{minipage}{0.48\textwidth}
  \begin{subfigure}{0.49\textwidth}
    \centering
    \caption*{Ours}
    \includegraphics[trim={212px 100px 287px 136px}, clip, width=\linewidth]{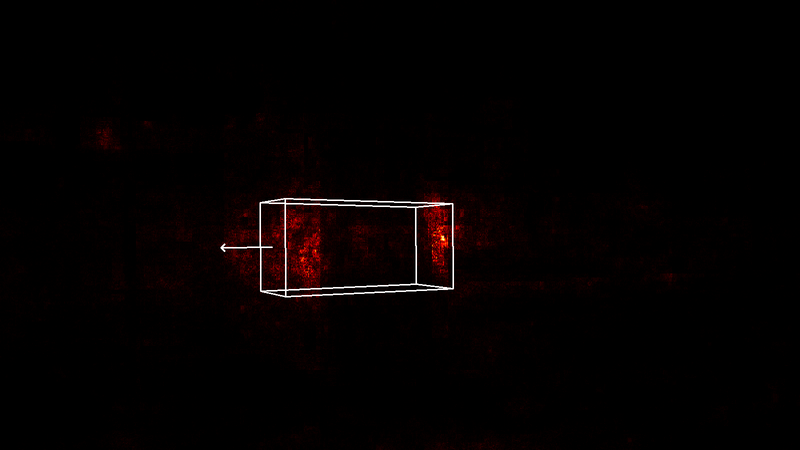}
  \end{subfigure}
  \begin{subfigure}{0.49\textwidth}
    \centering
    \caption*{BEVFusion~\cite{liu2022bevfusion}}
    \includegraphics[trim={212px 100px 287px 136px}, clip, width=\linewidth]{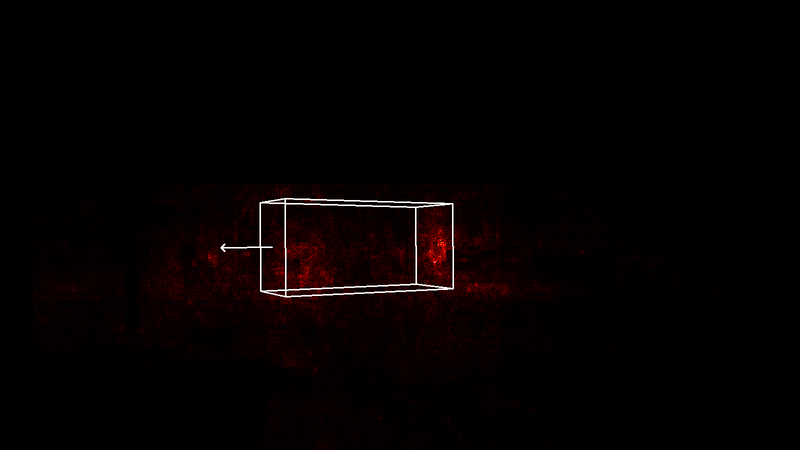}
  \end{subfigure}
  \\
  \begin{subfigure}{0.49\textwidth}
    \centering
    \includegraphics[trim={212px 100px 287px 136px}, clip, width=\linewidth]{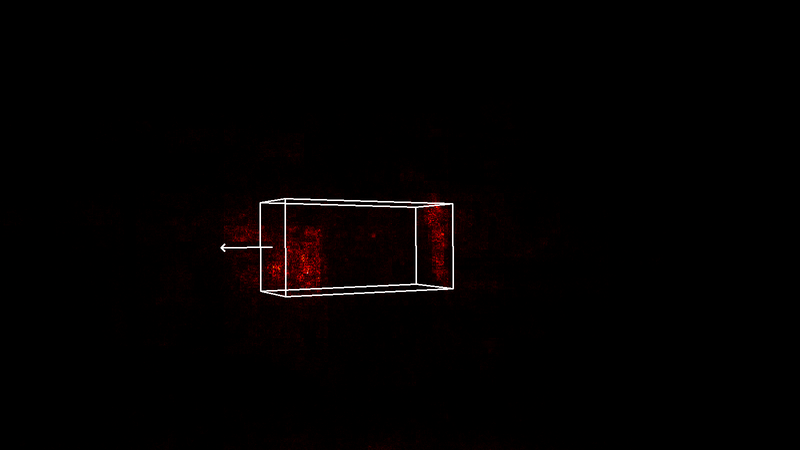}
  \end{subfigure}
  \begin{subfigure}{0.49\textwidth}
    \includegraphics[trim={212px 100px 287px 136px}, clip, width=\linewidth]{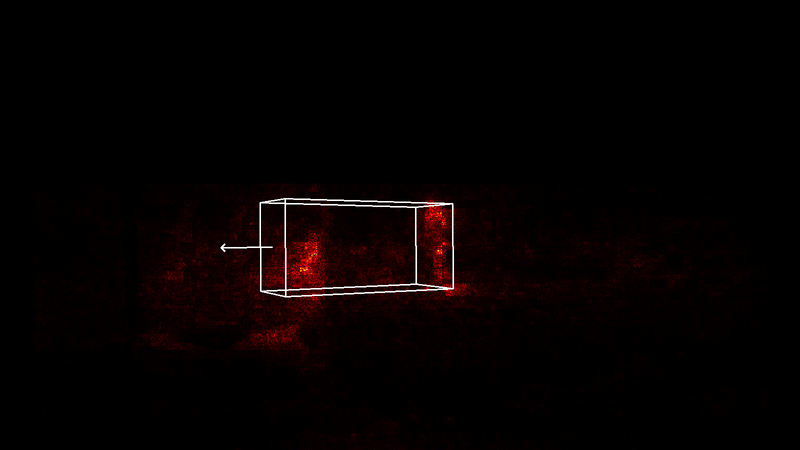}
  \end{subfigure}
  \\
  \begin{subfigure}{\textwidth}
  \centering
  \begin{tikzpicture}
    \node[anchor=south west,inner sep=0] at (0,0) {
    \includegraphics[width=0.6\linewidth]{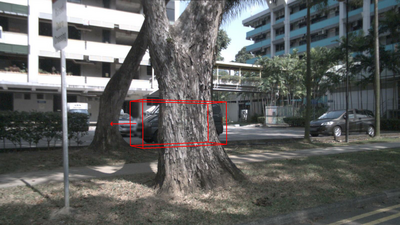}
    };
    \draw[white,dashed,thick] (0.16\linewidth,0.075\linewidth) rectangle (0.3843\linewidth, 0.2335\linewidth);
    \end{tikzpicture}
    \caption{}
  \end{subfigure}
\end{minipage}
\begin{minipage}{0.48\textwidth}
  \begin{subfigure}{0.49\textwidth}
    \centering
    \caption*{Ours}
    \includegraphics[trim={500px 100px 50px 171px}, clip, width=\linewidth]{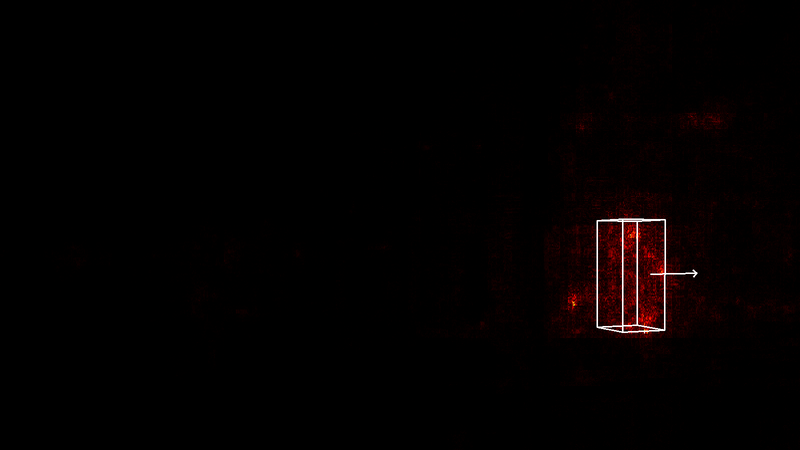}
  \end{subfigure}
  \begin{subfigure}{0.49\textwidth}
    \centering
    \caption*{BEVFusion~\cite{liu2022bevfusion}}
    \includegraphics[trim={500px 100px 50px 171px}, clip, width=\linewidth]{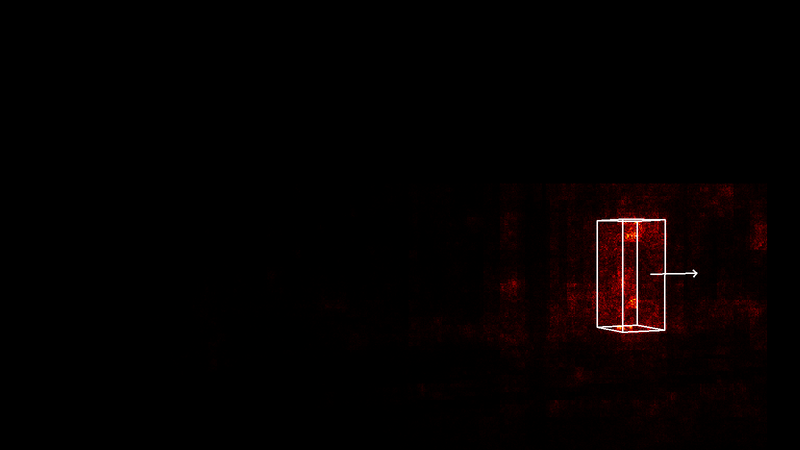}
  \end{subfigure}
  \\
  \begin{subfigure}{0.49\textwidth}
    \centering
    \includegraphics[trim={500px 100px 50px 171px}, clip, width=\linewidth]{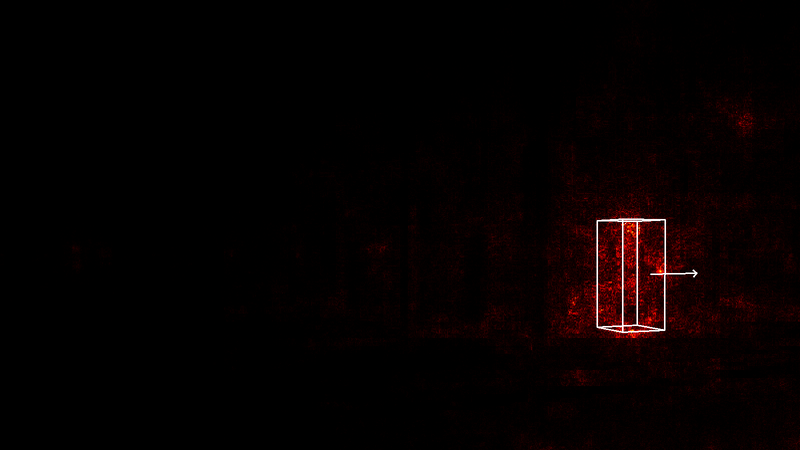}
  \end{subfigure}
  \begin{subfigure}{0.49\textwidth}
    \includegraphics[trim={500px 100px 50px 171px}, clip, width=\linewidth]{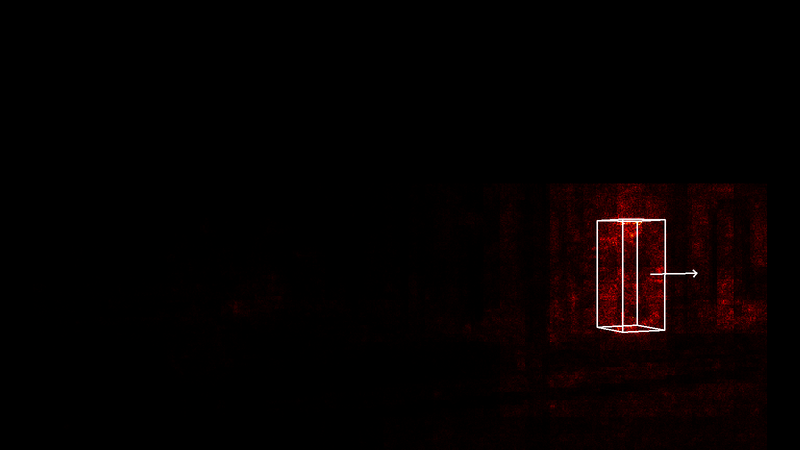}
  \end{subfigure}
  \\
  \begin{subfigure}{\textwidth}
  \centering
  \begin{tikzpicture}
    \node[anchor=south west,inner sep=0] at (0,0) {
    \includegraphics[width=0.6\linewidth]{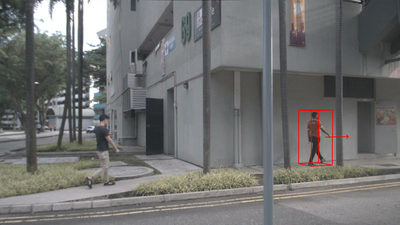}
    };
    \draw[white,dashed,thick] (0.375\linewidth,0.075\linewidth) rectangle (0.5625\linewidth, 0.208875\linewidth);
    \end{tikzpicture}
    \caption{}
  \end{subfigure}
\end{minipage}
\begin{minipage}{0.02\textwidth}
  \centering
  \begin{subfigure}{\textwidth}
    \caption*{\rotatebox{90}{\qquad \qquad \quad Cam+lidar}} 
  \end{subfigure}
  \\
  \begin{subfigure}{\textwidth}
    \caption*{\rotatebox{90}{\qquad \qquad \qquad \qquad \qquad Cam-only}} 
  \end{subfigure}
\end{minipage}
\end{center}
\caption{\label{fig:appendix:explainability}
\textbf{(a)} Similar attention pattern as highlighted in the main text for a close object which is well-represented by the lidar point cloud. \textbf{(b)} For the same object, but at a later frame when the car moved further away from the ego, our model attends to the same area when trained with camera and lidar as when trained with camera only (left). \textbf{(c)} For an occluded object, whose representation in the lidar point cloud is weaker, our model attends to the entire unoccluded area in both settings. BEVFusion~\cite{liu2022bevfusion} appears to consistently attend to a larger neighbourhood of pixels. \textbf{(d)} The pedestrian, who is not well-represented by the lidar point cloud, is fully attended by our model in the presence of lidar and camera.
}
\end{figure*}

\begin{figure*}
\begin{center}
  \begin{subfigure}{0.49\textwidth}
    \centering
    \caption*{Scene}
    \includegraphics[width=\linewidth]{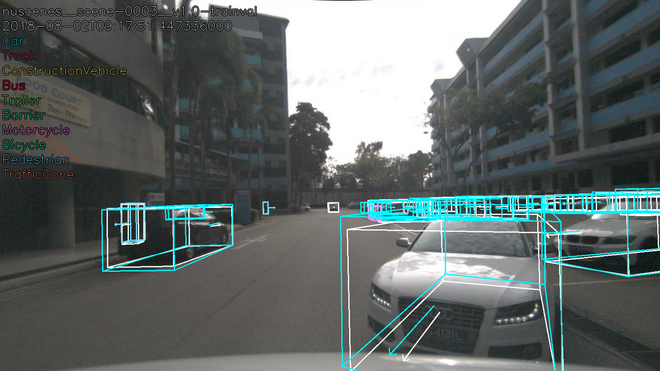}
    \caption{}
  \end{subfigure}
  \begin{subfigure}{0.49\textwidth}
    \centering
    \caption*{Mask}
    \includegraphics[width=\linewidth]{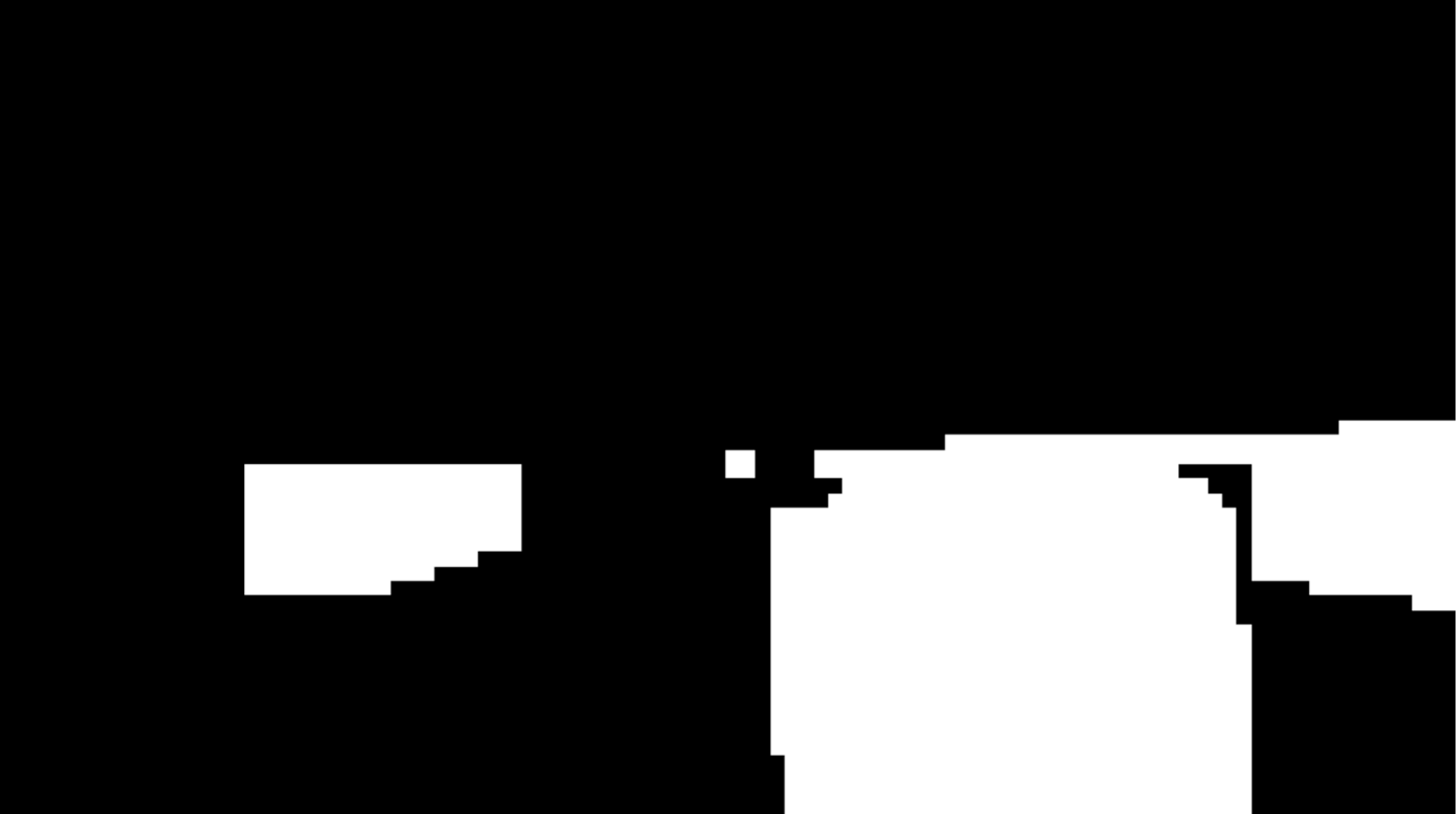}
    \caption{}
  \end{subfigure}
\end{center}
\caption{\label{fig:appendix:maskscene}
\textbf{(a)} Camera image $I_i$ with annotations highlighted in white and our model's predictions, in colour. \textbf{(b)} Binary image $\text{Mask}^{(i)}$ created by in-painting the annotations.  
}
\end{figure*}

\begin{figure*}
\begin{center}
  \begin{subfigure}{0.49\textwidth}
    \centering
    \caption*{Before fusion module}
    \includegraphics[width=0.9\linewidth]{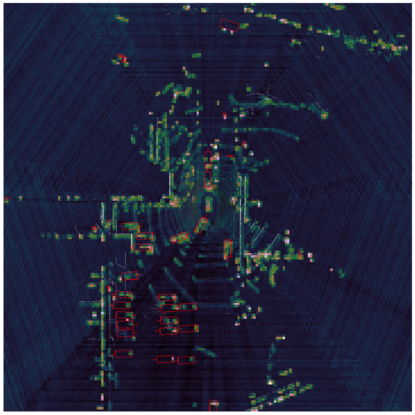}
    \caption{}
  \end{subfigure}
  \begin{subfigure}{0.49\textwidth}
    \centering
    \caption*{After fusion module}
    \includegraphics[width=0.9\linewidth]{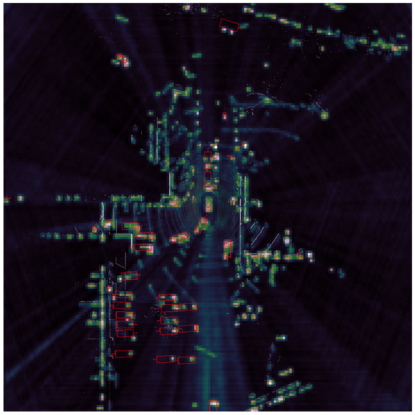}
    \caption{}
  \end{subfigure}
\end{center}
\caption{\label{fig:appendix:bev activations}
Activations in BEV space derived by summing up feature maps along the channel dimension. \textbf{(a)} uses the channel-wise concatenation of lidar and projected camera features. \textbf{(b)} uses the output of the fusion module, demonstrating its efficacy in suppressing background activations.
}
\end{figure*}

\begin{figure*}
\begin{center}
  \begin{subfigure}{0.49\textwidth}
    \centering
    \caption*{Ours}
    \includegraphics[width=0.9\linewidth]{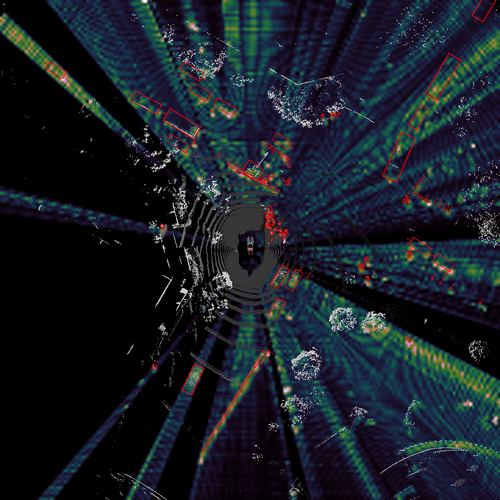}
    \caption{}
  \end{subfigure}
  \begin{subfigure}{0.49\textwidth}
    \centering
    \caption*{BEVFusion~\cite{liu2022bevfusion}}
    \includegraphics[width=0.9\linewidth]{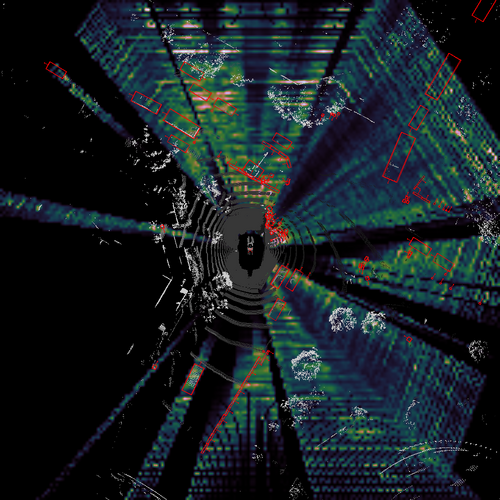}
    \caption{}
  \end{subfigure}
  \\
    \begin{subfigure}{0.49\textwidth}
    \centering
    \caption*{Ours}
    \includegraphics[width=0.9\linewidth]{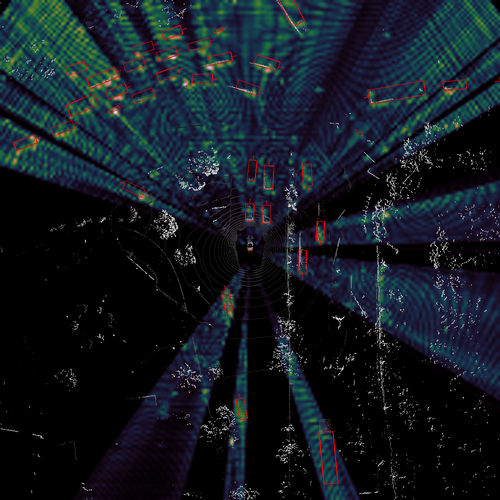}
    \caption{}
  \end{subfigure}
  \begin{subfigure}{0.49\textwidth}
    \centering
    \caption*{BEVFusion~\cite{liu2022bevfusion}}
    \includegraphics[width=0.9\linewidth]{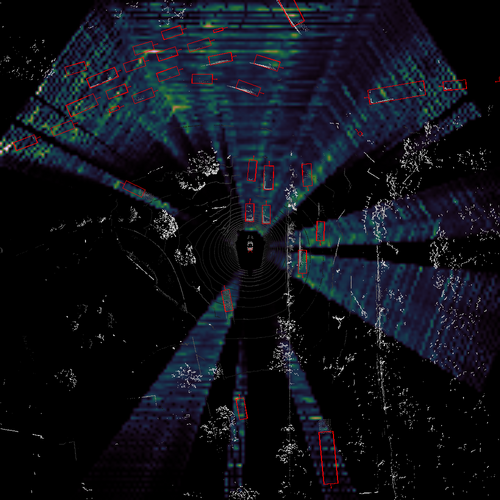}
    \caption{}
  \end{subfigure}
\end{center}
\caption{\label{fig:appendix:bev more examples}
Additional examples showcasing the weight of projected camera features onto the BEV space. All examples presented in the analysis are from the validation set.
}
\end{figure*}

To obtain \Cref{sec:explainability:bevlas} for our method, we first compute the full camera-to-BEV attention map $\text{Attn}^{(\text{cam}_i \rightarrow \text{bev})} \in \mathbb{R}^{H \times W \times N \times M}$. To do so, we extract the attention map of the last transformer decoder block by averaging over all heads, $\text{Attn}^{(\text{cam}_i \rightarrow \text{frustum})} \in \mathbb{R}^{H \times W \times D \times W'}$, where $(D \times W')$ corresponds to the frustum dimension. We construct $\text{Attn}^{(\text{cam}_i \rightarrow \text{bev})}$ by scattering the frustum attention values onto the BEV grid. Given camera image $I_i$, we then create $\text{Mask}^{(i)} \in \{0,1\}^{H \times W}$ by in-paint drawing the annotations, see~\cref{fig:appendix:maskscene}, and obtain the BEV attention $\text{Attn}^{(\text{bev})} \in \mathbb{R}^{N \times M}$ shown on \Cref{sec:explainability:bevlas} by projecting these camera features onto the BEV grid:
\begin{equation}
\text{Attn}^{(\text{bev})} = \max_{i, h, w} \text{Attn}_{h, w}^{(\text{cam}_i \rightarrow \text{bev})} \cdot \text{Mask}_{h, w}^{(i)}.
\end{equation}

To obtain a similar visualisation for the ``LiftSplat'' projection, see \Cref{sec:explainability:bevlss}, we adjust the implementation of~\cite{liu2022bevfusion} but use the same model weights. Firstly, we replace the feature map of image $I_i$ with $\text{Mask}^{(i)}$ and use that as input to the projection. This binary mask is ``lifted'' onto a 3D point cloud using the normalised depth classification weights $D_i$ for which we clipped the first 5 and last 5 depth bins.
$\text{Mask}^{(i)}$ thus acts as an indicator function and $D_i$ specifies the strength of correspondence between pixels and the 3D point cloud $P \in \mathbb{R}^{H \times W \times N_{D} \times 3}$. Secondly, during ``splatting'', we project points  onto the z = 0 plane and pool them using max. This operation ensures that the weight of attention for large objects in the final visualisation does not overpower that of smaller objects.

\subsection{Ensemble and test-time augmentations} \label{sec:suppl:ensemble}
For test-time-augmentation (TTA) and model ensembling, we use WBF~\cite{wbf} based on L2 distance metric per object category to decide which of the boxes to fuse. We first carry out TTA (using mirror and rotation augmentations) with WBF for each cell resolution, and then apply another WBF on the outputs from TTA of each model to get the final detections which we use for evaluations. For rotation augmentation, we use (-12.5, -6.25, 0, 6.25, 12.5) degrees.

\end{document}